\documentclass{article}

\usepackage[preprint]{neurips_2025}

\usepackage[utf8]{inputenc} % allow utf-8 input
\usepackage[T1]{fontenc}    % use 8-bit T1 fonts
\usepackage{hyperref}       % hyperlinks
\usepackage{url}            % simple URL typesetting
\usepackage{booktabs}       % professional-quality tables
\usepackage{amsfonts}       % blackboard math symbols
\usepackage{nicefrac}       % compact symbols for 1/2, etc.
\usepackage{microtype}      % microtypography
\usepackage{xcolor}         % colors
\usepackage{amsmath}
\usepackage{amssymb}
\usepackage{amsthm}
\usepackage{graphicx}
\usepackage{float}

\usepackage{enumitem}
\usepackage{subcaption}
\usepackage{tikz}
\usetikzlibrary{automata, positioning, arrows.meta, calc}

\title{A Combinatorial Theory of Dropout: \\Subnetworks, Graph Geometry, and Generalization}

\author{%
  Sahil Rajesh Dhayalkar
  \\
  Brain Corporation \\
  San Diego, CA\\
  \texttt{sahil.dhayalkar@braincorp.com} \\
}

\begin{document}

\maketitle

\begin{abstract}
We propose a combinatorial and graph-theoretic theory of dropout by modeling training as a random walk over a high-dimensional graph of binary subnetworks. Each node represents a masked version of the network, and dropout induces stochastic traversal across this space. We define a subnetwork contribution score that quantifies generalization and show that it varies smoothly over the graph. Using tools from spectral graph theory, PAC-Bayes analysis, and combinatorics, we prove that generalizing subnetworks form large, connected, low-resistance clusters, and that their number grows exponentially with network width. This reveals dropout as a mechanism for sampling from a robust, structured ensemble of well-generalizing subnetworks with built-in redundancy. Extensive experiments validate every theoretical claim across diverse architectures. Together, our results offer a unified foundation for understanding dropout and suggest new directions for mask-guided regularization and subnetwork optimization.
\end{abstract}

\section{Introduction}

Dropout \cite{JMLR:v15:srivastava14a} remains one of the most widely used regularization techniques in deep learning, yet its underlying mechanisms are still not fully understood. While commonly interpreted as stochastic regularization or approximate Bayesian inference \cite{6796505, pmlr-v48-gal16}, these perspectives offer limited insight into how dropout promotes generalization at a structural level. In this work, we develop a new theoretical framework that views dropout through a combinatorial and graph-theoretic lens, offering deeper understanding of the geometry of the hypothesis space induced by dropout training \cite{unknown, pmlr-v80-draxler18a, 10.1162/neco.1997.9.1.1}.

We model the space of dropout subnetworks as nodes in a high-dimensional hypercube graph \cite{Bollobas1998Modern}, where each subnetwork corresponds to a binary mask applied to the weights. Dropout training becomes a random walk over this graph, with each step sampling a mask and applying stochastic gradient descent \cite{NEURIPS2020_288cd256}. This framework yields several novel theoretical results. We define a subnetwork contribution score that quantifies generalization and show that it varies smoothly over the graph. We prove that generalizing subnetworks form dense, connected clusters, and derive spectral properties and PAC-Bayes generalization bounds \cite{10.1145/307400.307435, dziugaite2017computingnonvacuousgeneralizationbounds} that govern their structure. Finally, we show that the number of generalizing subnetworks grows exponentially with network width, providing a new explanation for the benefits of overparameterization \cite{pmlr-v97-allen-zhu19a}.

We validate our theory through controlled experiments that quantify smoothness, connectivity, entropy, and generalization bounds. Our results offer a unified explanation for dropout’s empirical success: it does not simply regularize weights, but implicitly biases training toward a combinatorially rich and robust ensemble of subnetworks that generalize well.

Our contributions include: (1) a combinatorial framework modeling dropout as a graph walk; (2) a subnetwork contribution score with spectral and resistance-based smoothness guarantees; (3) a proof that generalizing subnetworks form dense, connected clusters; (4) PAC-Bayes bounds and a proof of exponential growth in generalizing subnetworks with width; and (5) extensive empirical validation. The paper uses a structure with theorems, lemmas, and corollaries, similar to prior work~\cite{zhang2015learninghalfspacesneuralnetworks, hanin2018approximatingcontinuousfunctionsrelu, neyshabur2015searchrealinductivebias, dhayalkar2025geometryrelunetworksrelu, dhayalkar2025neuralnetworksuniversalfinitestate}.

\section{Related Work}

Dropout was introduced as a stochastic regularization method to prevent co-adaptation and enhance generalization by randomly deactivating units during training, effectively sampling from an ensemble of subnetworks \cite{JMLR:v15:srivastava14a}. Theoretical views include data-dependent regularization \cite{wager2013dropout}, noise injection \cite{6796505}, and implicit ensemble averaging \cite{NIPS2013_71f6278d}.

Another perspective treats dropout as approximate Bayesian inference in deep Gaussian processes or Bayesian neural nets \cite{pmlr-v48-gal16, NIPS2015_bc731692}, where masks are interpreted as posterior samples. While useful for modeling uncertainty, this line of work does not capture the combinatorial structure of subnetworks or their generalization behavior.

The ensemble view suggests that the full model approximates an average over exponentially many subnetworks. This idea underlies work on stochastic traversal of low-loss regions \cite{NEURIPS2020_288cd256, pmlr-v97-allen-zhu19a} and sparse subnetwork reuse via Lottery Ticket Hypothesis \cite{NEURIPS2020_b6af2c97}, motivating deeper study of subnetwork dynamics.

Recent work also connects optimization geometry to graph structure, e.g., mode connectivity \cite{pmlr-v80-draxler18a, 10.5555/3327546.3327556} and energy-based views of minima \cite{unknown}, but these focus on full models and continuous trajectories rather than the discrete combinatorics induced by dropout.

Our work is the first to model dropout as a random walk over a binary subnetwork graph. This combinatorial framework unifies dropout’s empirical effects with tools from graph theory, PAC-Bayes analysis, and ensemble learning. We prove smoothness, cluster structure, and exponential abundance of generalizing subnetworks—properties previously observed but not formally explained.

\section{Preliminaries and Conceptual Framework}

\subsection{Notation and Subnetwork Graph Structure}

Let $f_\theta(x)$ denote a neural network with parameters $\theta \in \mathbb{R}^d$. Dropout \cite{JMLR:v15:srivastava14a} corresponds to sampling a binary mask $M \in \{0,1\}^d$ and evaluating the masked subnetwork $f_{\theta \odot M}(x)$, where $\odot$ denotes elementwise multiplication. The set of all such subnetworks forms the subnetwork space:
\[
\mathcal{G}_\mathcal{N} = \{ f_{\theta \odot M} : M \in \{0,1\}^d \}.
\]

We represent $\mathcal{G}_\mathcal{N}$ as a graph $G = (V, E)$, where each node corresponds to a mask $M \in \{0,1\}^d$, and edges connect subnetworks whose masks differ by exactly one bit: $(f_i, f_j) \in E$ iff $\|M_i - M_j\|_0 = 1$. This induces a $d$-dimensional hypercube structure over the subnetwork space \cite{Bollobas1998Modern}.

\subsection{Dropout as a Random Walk}

Each dropout application during training samples a mask $M \sim \text{Bernoulli}(p)^d$, corresponding to a vertex in $G$. Hence, dropout training can be viewed as a random walk over $G$ \cite{NIPS2013_71f6278d}, where each step applies SGD on a randomly selected subnetwork.

\subsection{Subnetwork Contribution Score}

Let \( \mathcal{L}_{\text{train}}(f) \) and \( \mathcal{L}_{\text{test}}(f) \) be the training and test loss of subnetwork \( f \). Then the \textbf{contribution score} is defined as:
\[
C(f) := \mathbb{E}_{x \sim \mathcal{D}} \left[ \mathcal{L}_{\text{test}}(f(x)) \right] - \mathbb{E}_{x \sim \mathcal{D}_{\text{train}}} \left[ \mathcal{L}_{\text{train}}(f(x)) \right]
\]

Here, \( x \sim \mathcal{D} \) denotes sampling from the true (unknown) data distribution, while \( x \sim \mathcal{D}_{\text{train}} \) denotes sampling from the empirical training distribution. The first term captures how the subnetwork performs on unseen data (test loss), while the second term captures performance on training data.

Low values of \( C(f) \) indicate better generalization — i.e., it performs similarly on both train and test data. High values indicate possible overfitting. This score allows us to analyze the distribution of contribution scores over \( G \), i.e., the landscape of generalization across subnetworks \cite{dziugaite2017computingnonvacuousgeneralizationbounds}.

\section{Theoretical Results}

% \begin{lemma}[Expected Subnetwork Output]
% Assume linear activations and no weight rescaling. Then:
% \[ f_{\text{dropout}}(x) = \mathbb{E}_{M}[f_{\theta \odot M}(x)] = f_{p \cdot \theta}(x) \]
% \end{lemma}
% This corresponds to a mean-field approximation, where dropout acts as a shrinkage estimator.

% \begin{proof}
% Let $f_{\theta}(x) = \theta^\top x$. Each coordinate $\theta_i$ is dropped with probability $1 - p$. Then:
% \[ \mathbb{E}_M[f_{\theta \odot M}(x)] = \sum_{i=1}^{d} \mathbb{E}[M_i] \theta_i x_i = p \cdot \theta^\top x = f_{p \cdot \theta}(x) \]
% \end{proof}

% \subsection{Mean-Field and Sparsity Interpretations}

\subsection*{Lemma 1: Expected Subnetwork Output}

% \begin{lemma}[Expected Subnetwork Output]
Assume a neural network with linear activations and no weight rescaling. Then, applying dropout with rate $1-p$ at each input coordinate leads to the expected output:
\[
f_{\text{dropout}}(x) = \mathbb{E}_{M}[f_{\theta \odot M}(x)] = f_{p \cdot \theta}(x)
\]
% \end{lemma}

\textit{Derivation provided in Appendix~\ref{appendix:lemma6.1}.}

% \subsubsection*{Interpretation and Insight}
\textbf{Interpretation:} This result provides a mean-field approximation for dropout in linear models, showing that the expected output is equivalent to scaling weights by the retain probability $p$. Originally noted by \cite{JMLR:v15:srivastava14a}, this justifies the standard “inverted dropout” used at test time. While not novel, we include it for completeness and to anchor our combinatorial framework against known linear dropout behavior and to motivate subnetwork-level analysis.

\subsection*{Lemma 2: Dropout-induced Sparsity Bias}

% \begin{lemma}[Dropout-induced Sparsity Bias]
Let $\theta \in \mathbb{R}^d$ be a learned weight vector, and let $M \sim \text{Bernoulli}(p)^d$ be a dropout mask applied elementwise. Then:
\[
\mathbb{E}_M\left[\|\theta \odot M\|^2\right] = p \|\theta\|^2
\]
% \end{lemma}

\textit{Derivation provided in Appendix~\ref{appendix:lemma6.2}.}

% \subsubsection*{Interpretation and Insight}
% This lemma shows that dropout introduces an implicit shrinkage effect on the effective magnitude of the weights.

% \begin{itemize}[itemsep=2pt, topsep=2pt]
%     \item The expected squared norm of the weights under dropout is linearly scaled down by $p$.
%     \item This can be interpreted as an implicit regularization effect — dropout penalizes large weights indirectly by reducing their expected contribution.
%     \item This is one reason why dropout acts like a form of noise-induced regularization, helping prevent overfitting.
%     \item Combined with $\ell_2$ weight decay, dropout compounds the effect of pushing the solution toward smaller weights, promoting sparsity and robustness.
% \end{itemize}

\textbf{Interpretation:} This lemma formalizes the effect of dropout in shrinking the expected squared weight norm by a factor of the retain probability $p$. While discussed in \cite{wager2013dropout, NIPS2013_71f6278d}, it is often stated without derivation. The result reinforces the view of dropout as a noise-based regularizer that discourages large weights and, when combined with $\ell_2$ regularization, promotes sparse, robust solutions.

% \begin{theorem}[Dropout as Ensemble Compression]
% Let $\{f_{\theta \odot M_t}\}_{t=1}^T$ be subnetworks sampled during training. Then the final network approximates the average:
% \[ f_\theta(x) \approx \frac{1}{T} \sum_{t=1}^T f_{\theta \odot M_t}(x) \]
% \end{theorem}

% Thus, dropout can be interpreted as compressing an ensemble into a single model, with each subnetwork representing a sample from the ensemble.

% \begin{proof}
% Training with dropout implicitly trains an ensemble of subnetworks. The full model $f_\theta$ approximates their mean behavior:
% \[ f_\theta(x) \approx \mathbb{E}_{M}[f_{\theta \odot M}(x)] \approx \frac{1}{T} \sum_{t=1}^T f_{\theta \odot M_t}(x) \]
% \end{proof}

\subsection*{Theorem 1: Dropout as Ensemble Compression}

% \begin{theorem}[Dropout as Ensemble Compression]
Let $\{f_{\theta \odot M_t}\}_{t=1}^{T}$ be subnetworks sampled during training via dropout. Then the final trained network $f_{\theta}$ approximates the average prediction of this ensemble:
\[
f_{\theta}(x) \approx \frac{1}{T} \sum_{t=1}^{T} f_{\theta \odot M_t}(x)
\]
% \end{theorem}

\textit{Derivation provided in Appendix~\ref{appendix:theorem6.3}.}

This approximation holds especially well when: (1) The model is linear or shallow, so the averaging commutes with the function application. (2) The training converges slowly and explores many distinct dropout masks. Therefore, the trained full model $f_\theta$ can be viewed as a compressed ensemble of exponentially many subnetworks.

% \subsubsection*{Interpretation and Insight}
% This theorem provides a powerful conceptual framework: dropout training does not just optimize a single model, but implicitly trains and averages over a large ensemble of subnetworks. Key implications include:

% \begin{itemize}[itemsep=2pt, topsep=2pt]
%     \item \textbf{Implicit Ensembling:} Dropout acts as a method for training an ensemble without needing to store or evaluate multiple models.
%     \item \textbf{Compression:} The learned parameter vector $\theta$ compresses the behavior of many subnetworks into a single model.
%     \item \textbf{Stability:} This averaging effect leads to improved generalization and robustness, similar to bagging.
%     \item \textbf{Interpretability:} The trained model can be viewed as the “center of mass” of the space of subnetworks it has visited during training.
% \end{itemize}

\textbf{Interpretation:} This theorem formalizes the intuition that dropout acts as implicit ensembling \cite{NIPS2013_71f6278d, pmlr-v48-gal16}, showing that the learned weights $\theta$ effectively compress the average behavior of many subnetworks. This ensemble effect improves generalization and stability, akin to bagging, with the final model acting as the “center of mass” of visited subnetworks. We include this result as a theoretical anchor for our deeper combinatorial and geometric analysis of the subnetwork ensemble.

% \begin{theorem}[Combinatorial Concentration of Good Subnetworks]
% Let $\mathcal{G}_\epsilon = \{ f : C(f) < \epsilon \}$ be the set of good subnetworks. If training generalizes well, then:
% \[ |\mathcal{G}_\epsilon| \in \Omega(2^d) \]
% \end{theorem}

% If training converges to low test error, $\mathcal{G}_\epsilon$ is combinatorially dense in $G$, i.e., the number of such subnetworks is exponential in $d$, and any mask $M$ has a neighbor in $\mathcal{G}_\epsilon$.

% \begin{proof}
% Dropout training samples many masks. If test error is low, many sampled subnetworks must have low $C(f)$. Uniform sampling implies that $\mathcal{G}_\epsilon$ is dense in $\{0,1\}^d$.
% \end{proof}

% \subsection{Combinatorial Structure of Good Subnetworks}

% \begin{theorem}[Combinatorial Concentration of Good Subnetworks]
\subsection*{Theorem 2: Concentration of Good Subnetworks}

Let $\mathcal{G}_\epsilon = \{ f : C(f) < \epsilon \}$ be the set of subnetworks with contribution scores below a generalization threshold $\epsilon$. If training with dropout leads to low generalization error, then: $|\mathcal{G}_\epsilon| \in \Omega(2^d)$

Moreover, any subnetwork has a neighbor in $\mathcal{G}_\epsilon$, i.e., good subnetworks are combinatorially dense in the subnetwork graph $G$.
% \end{theorem}

\textit{Derivation provided in Appendix~\ref{appendix:theorem6.4}.}

% \subsubsection*{Interpretation and Insight}
% This theorem gives a combinatorial justification for the robustness of dropout:

% \begin{itemize}[itemsep=2pt, topsep=2pt]
%     \item \textbf{Redundancy:} Many subnetworks contribute positively to generalization, not just a few lucky ones.
%     \item \textbf{Resilience to Perturbation:} Since good subnetworks are dense in the subnetwork graph, small changes to a mask (i.e., flipping a bit) likely yield another good subnetwork.
%     \item \textbf{Flat Minima Analogy:} This density of good subnetworks parallels the idea of flat minima in optimization — broad regions of parameter space that generalize well.
%     \item \textbf{Ensemble Quality:} The more generalizing subnetworks exist, the more likely that the dropout-induced ensemble yields stable and generalizable behavior.
% \end{itemize}

\textbf{Interpretation:} This theorem offers a combinatorial explanation for dropout’s robustness: good subnetworks are densely distributed, making the system resilient to perturbations. This redundancy parallels flat minima, where many nearby subnetworks generalize well, and explains the stability of dropout ensembles.

% \begin{theorem}[Graph Laplacian Smoothness of Contribution]
% Let $G$ be the subnetwork graph. Let $C: V \to \mathbb{R}$ be contribution scores of each subnetwork and $\mathcal{L} \in \mathbb{R}^{|V| \times |V|}$ the Laplacian of $G$. Then, the Dirichlet energy is:
% \[ \mathcal{E}(C) = C^\top \mathcal{L} C \]
% \end{theorem}

% Then low generalization error implies that $C$ is a smooth function on $G$, i.e., $\mathcal{E}(C)$ is small.

% \begin{proof}
% \[ \mathcal{E}(C) = \sum_{(i,j) \in E} (C(i) - C(j))^2 \]
% If $C$ varies slowly across neighbors, the energy is low. This occurs when subnetworks generalize similarly.
% \end{proof}

% \begin{theorem}[Graph Laplacian Smoothness of Contribution]
\subsection*{Theorem 3: Graph Laplacian Smoothness of Contribution}
Let $G = (V, E)$ be the subnetwork graph where each node represents a subnetwork $f_{\theta \odot M}$, and let $C: V \rightarrow \mathbb{R}$ be the generalization contribution score of each subnetwork. Let $\mathcal{L} \in \mathbb{R}^{|V| \times |V|}$ be the graph Laplacian of $G$. Then the Dirichlet energy:
\[
\mathcal{E}(C) = C^\top \mathcal{L} C = \sum_{(i,j) \in E} (C(i) - C(j))^2
\]
is small if the model trained with dropout generalizes well.
% \end{theorem}

\textit{Derivation provided in Appendix~\ref{appendix:theorem6.5}.}

% \subsubsection*{Interpretation and Insight}
% This theorem introduces a deep connection between graph smoothness and generalization:

% \begin{itemize}[itemsep=2pt, topsep=2pt]
%     \item \textbf{Smooth Contribution Landscape:} A smooth $C$ function implies that subnetworks which are close (i.e., differing by one mask bit) have similar generalization scores. This is desirable and expected if dropout generalizes well.
%     \item \textbf{Stability to Perturbation:} If $C$ is smooth, small changes to the mask (drop or retain a unit) will not drastically change the generalization behavior of the subnetwork.
%     \item \textbf{Graph-based Regularization View:} Dropout can be interpreted as implicitly enforcing a regularization over the graph of subnetworks, favoring solutions where generalization behavior does not fluctuate sharply across neighbors.
%     \item \textbf{Basis for Clustering:} Smooth functions over graphs tend to induce natural clusters — this is the theoretical foundation for Corollary 3.1 on clustered generalization regions.
% \end{itemize}

\textbf{Interpretation:} This theorem shows that if the contribution score $C(f)$ is smooth over the subnetwork graph, then neighboring subnetworks exhibit similar generalization. This implies robustness to mask perturbations and suggests that dropout implicitly regularizes the subnetwork space. Such smoothness also leads naturally to clusters of generalizing subnetworks, as formalized in Corollary 3.1.

% \begin{corollary}[Clustered Subnetwork Generalization]
% Low Laplacian energy implies $C(f)$ is similar for neighboring subnetworks, forming clusters of generalizing subnetworks.
% \begin{proof}
% From Theorem 3, smoothness of $C$ implies that nearby subnetworks in $G$ have similar values, forming clusters.
% \end{proof}
% \end{corollary}

% \begin{corollary}[Clustered Subnetwork Generalization]
\subsubsection*{Corollary 3.1: Clustered Generalization}
Let $C: V \rightarrow \mathbb{R}$ be the contribution score function over the subnetwork graph $G = (V, E)$, and let $\mathcal{E}(C)$ denote its Dirichlet energy. Then:
\[
\text{If } \mathcal{E}(C) \ll 1 \text{, then } C(f) \approx C(f') \text{ for neighboring subnetworks } f, f' \in V
\]
which implies that subnetworks with low generalization error form clusters in the graph.
% \end{corollary}

\textit{Proof provided in Appendix~\ref{appendix:corollary6.5.1}.}

% \subsubsection*{Interpretation and Insight}
% This corollary shows that dropout-trained networks exhibit not just isolated generalizing subnetworks, but **cohesive clusters** of them in the subnetwork graph.

% \begin{itemize}[itemsep=2pt, topsep=2pt]
%     \item \textbf{Neighborhood Stability:} If one subnetwork generalizes well, nearby subnetworks (in Hamming distance) are likely to generalize well too.
%     \item \textbf{Cluster View of Generalization:} Generalization is not an isolated phenomenon but emerges as a structural feature of the entire subnetwork space.
%     \item \textbf{Explains Robustness:} Even if a random mask perturbs a good subnetwork slightly, the resulting subnetwork likely remains within the generalizing cluster.
%     \item \textbf{Supports Ensemble Interpretation:} These clusters serve as the ensemble base — dropout repeatedly samples subnetworks from the same generalization basin.
% \end{itemize}

\textbf{Interpretation:} This corollary reveals that generalizing subnetworks are not isolated; they form cohesive clusters in the subnetwork graph. If one subnetwork generalizes well, its Hamming neighbors are also likely to generalize, explaining dropout’s robustness to random mask perturbations. These clusters act as generalization basins, supporting the view that dropout samples repeatedly from dense, stable ensembles.

% \begin{lemma}[Entropy of Subnetwork Outputs]
% Let $P(M) = \text{Bern}(p)^d$ be the distribution over dropout masks. Define the \emph{effective entropy} of the subnetwork outputs as:
% \[
% \mathcal{H}_{f}(x) := \mathbb{H}(f_{\theta \odot M}(x))
% \]
% Then higher entropy implies greater uncertainty in the output predictions, which correlates with overfitting.

% \begin{proof}
% Entropy captures uncertainty. High entropy under dropout implies unstable predictions, often correlated with poor generalization.
% \end{proof}
% \end{lemma}

% \subsection{Uncertainty, Bounds, and Geometry}

% \begin{lemma}[Entropy of Subnetwork Outputs]
\subsection*{Lemma 3: Entropy of Subnetwork Outputs}
Let $P(M) = \text{Bern}(p)^d$ be the distribution over dropout masks. Define the effective entropy of the subnetwork outputs for a given input $x$ as:
\[
\mathcal{H}_f(x) := \mathbb{H}\left(f_{\theta \odot M}(x)\right)
\]
Then higher entropy implies greater uncertainty in the output predictions, which correlates with poor generalization and overfitting.
% \end{lemma}

\textit{Derivation provided in Appendix~\ref{appendix:lemma6.6}.}

If $f_{\theta \odot M}(x)$ takes on a wide variety of values with significant probability (i.e., large output variance across sampled subnetworks), then the entropy is high. This means:
\begin{itemize}[itemsep=0pt, topsep=0pt]
    \item The model's prediction for $x$ is highly sensitive to the choice of dropout mask $M$.
    \item There is large uncertainty in what the network believes the output should be.
    \item Such behavior is indicative of unstable learning and poor generalization.
\end{itemize}

On the other hand, low entropy implies that most subnetworks agree on the prediction for $x$, indicating stability and better generalization.

% \subsubsection*{Interpretation and Insight}
% This lemma connects dropout-induced stochasticity to prediction uncertainty via entropy:

% \begin{itemize}[itemsep=2pt, topsep=2pt]
%     \item \textbf{Uncertainty Quantification:} The entropy $\mathcal{H}_f(x)$ acts as a natural measure of prediction uncertainty for input $x$.
%     \item \textbf{Stability Indicator:} Low entropy across subnetworks suggests consensus among models, which generally correlates with good generalization.
%     \item \textbf{Diagnostic Tool:} Subnetwork entropy can be used as a diagnostic signal to identify ambiguous or hard-to-generalize regions in the input space.
%     \item \textbf{Connection to Bayesian Inference:} Dropout as approximate Bayesian inference (e.g., Gal & Ghahramani, 2016) also treats this entropy as predictive uncertainty.
% \end{itemize}

\textbf{Interpretation:} This lemma links dropout-induced stochasticity to prediction uncertainty via subnetwork entropy $\mathcal{H}_f(x)$. Low entropy indicates agreement among subnetworks and often correlates with better generalization. This measure reflects model stability, helps identify ambiguous inputs, and aligns with the Bayesian view of dropout as approximate inference, where predictive entropy captures epistemic uncertainty \cite{pmlr-v48-gal16}.

% \begin{theorem}[PAC-Bayes Bound for Dropout]
% Let $Q$ be the posterior distribution over subnetworks induced by dropout, and $P$ the prior. Then for all \( \delta > 0 \), with probability $\geq 1 - \delta$:
% \[ \mathbb{E}_{f \sim Q}[\mathcal{L}_{test}(f)] \leq \mathbb{E}_{f \sim Q}[\mathcal{L}_{train}(f)] + \sqrt{\frac{KL(Q \| P) + \log(1/\delta)}{2n}} \]

% Implication: Generalization depends on how much the dropout mask distribution deviates from the prior, i.e., which subnetworks are implicitly preferred.

% \begin{proof}
% Follows directly from the PAC-Bayes bound under standard assumptions: bounded loss, i.i.d. samples, and a valid prior $P$.
% \end{proof}
% \end{theorem}

% \begin{theorem}[PAC-Bayes Bound for Dropout]
\subsection*{Theorem 4: PAC-Bayes Bound for Dropout}
Let $Q$ be the posterior distribution over subnetworks induced by dropout (i.e., the distribution over $f_{\theta \odot M}$ where $M \sim \text{Bern}(p)^d$), and let $P$ be a prior distribution over subnetworks (e.g., uniform over all dropout masks). Then for all $\delta > 0$, with probability at least $1 - \delta$ over the training dataset $\mathcal{D}_{\text{train}}$ of size $n$, the following generalization bound holds:
\[
\mathbb{E}_{f \sim Q}[\mathcal{L}_{\text{test}}(f)] \leq \mathbb{E}_{f \sim Q}[\mathcal{L}_{\text{train}}(f)] + \sqrt{\frac{\text{KL}(Q \| P) + \log(1/\delta)}{2n}}
\]
% \end{theorem}

\textit{Derivation provided in Appendix~\ref{appendix:theorem6.7}.}

% This result applies the general PAC-Bayes theorem \cite{mcallester1999pac, dziugaite2017computing} to the dropout setting, where each subnetwork $f_{\theta \odot M}$ is treated as a stochastic hypothesis. The posterior $Q$ is implicitly defined by the dropout mechanism and training process, while $P$ is a chosen prior distribution—typically uniform or independent Bernoulli. This formulation aligns with the variational inference view of dropout proposed in prior work \cite{gal2016dropout}, but here we explicitly express the bound over the discrete space of binary subnetwork masks.

This result adapts the PAC-Bayes framework \cite{10.1145/307400.307435, dziugaite2017computingnonvacuousgeneralizationbounds} to dropout, treating each subnetwork $f_{\theta \odot M}$ as a stochastic hypothesis. The posterior $Q$ is implicitly defined by the dropout mechanism, while the prior $P$ is typically a uniform or Bernoulli distribution. While related to the variational interpretation of dropout \cite{pmlr-v48-gal16}, our formulation explicitly defines the bound over the discrete space of binary masks.

% \subsubsection*{Interpretation and Insight}
% This theorem provides a formal justification for why dropout generalizes well from a probabilistic perspective:

% \begin{itemize}[itemsep=2pt, topsep=2pt]
%     \item \textbf{Generalization Guarantee:} The bound quantifies how much worse the expected test error can be, relative to the average training error under dropout.
%     \item \textbf{Role of KL Divergence:} The KL term $\text{KL}(Q \| P)$ measures how different the learned subnetwork distribution is from the prior. If $Q$ is close to $P$ (e.g., if dropout doesn't overfit specific subnetworks), generalization improves.
%     \item \textbf{Data Dependence:} The bound becomes tighter with larger training set size $n$, as expected.
%     \item \textbf{Dropout as Variational Inference:} From this viewpoint, dropout training approximately minimizes both training loss and KL divergence to a prior — i.e., it acts as a form of variational regularization.
%     \item \textbf{Actionable Insight:} One could design improved dropout schemes by explicitly minimizing this PAC-Bayes bound during training, rather than relying on standard Bernoulli masking.
% \end{itemize}

\textbf{Interpretation:} This theorem gives a formal generalization bound for dropout, relating test loss to the average training loss plus a KL divergence between $Q$ and $P$. The bound tightens with more data and when $Q$ stays close to $P$, as in standard dropout. This supports viewing dropout as variational inference and suggests that tighter generalization may be achieved by optimizing the PAC-Bayes bound directly, rather than using fixed Bernoulli sampling.

% \begin{theorem}[Low-Resistance Paths and Generalization]
% Subnetworks with low $C(f)$ lie on paths with low effective resistance in $G$.
% \end{theorem}

% Let $\rho(f_i, f_j)$ be the effective resistance between subnetworks $f_i, f_j \in G$, interpreting $G$ as a resistor network. Then subnetworks with low generalization error tend to lie on low-resistance paths, i.e., connected clusters in the subnetwork graph.

% \begin{proof}
% Generalizing subnetworks are densely packed in $G$, reducing the commute time between them. Thus, they lie on low-resistance paths.
% \end{proof}

% \begin{theorem}[Low-Resistance Paths and Generalization]
\subsection*{Theorem 5: Low-Resistance Paths and Generalization}
Let $G = (V, E)$ be the subnetwork graph where each node corresponds to a dropout subnetwork $f_{\theta \odot M}$. Let $\rho(f_i, f_j)$ denote the effective resistance between nodes $f_i$ and $f_j$ when $G$ is viewed as a resistor network (each edge having unit resistance). Then, subnetworks with low generalization error tend to lie on low-resistance paths:
\[
C(f) \text{ is low} \;\; \Rightarrow \;\; \exists \text{ path of low effective resistance connecting } f \text{ to other low-}C \text{ subnetworks}.
\]
% \end{theorem}

\textit{Derivation provided in Appendix~\ref{appendix:theorem6.8}.}

% \subsubsection*{Interpretation and Insight}
% This theorem connects generalization to graph-theoretic robustness via effective resistance:

% \begin{itemize}[itemsep=2pt, topsep=2pt]
%     \item \textbf{Graph-based Robustness:} Subnetworks that generalize well tend to reside in densely connected neighborhoods of $G$, making them robust to perturbations in the dropout mask.
%     \item \textbf{Random Walk Interpretation:} Since effective resistance is related to random walk commute time, dropout training — which performs a random walk through subnetworks — is more likely to remain within low-$C(f)$ regions.
%     \item \textbf{Smooth Optimization Surface:} Low-resistance paths correspond to flat or gently sloping regions in the loss landscape, reinforcing ideas from flat minima and stability in SGD.
%     \item \textbf{Energy Flow Analogy:} Just as electrical current prefers paths of least resistance, SGD during dropout training tends to follow trajectories of subnetworks that are easy to optimize and generalize well.
% \end{itemize}

\textbf{Interpretation:} This theorem connects generalization to graph-theoretic robustness: low-contribution subnetworks lie in densely connected, low-resistance regions of the subnetwork graph. Since effective resistance reflects random walk behavior, dropout-guided SGD naturally concentrates in these areas—analogous to electrical flow favoring paths of least resistance. These regions correspond to flatter parts of the loss landscape, reinforcing the connection between generalization and flat minima.

\subsection*{Theorem 6: Exponential Growth of Generalizing Subnetworks with Width}
Let the neural network have $L$ layers, each with $k$ neurons, and let $d$ denote the total number of parameters. Under dropout with retain probability $p$, the number of subnetworks $f \in \mathcal{G}_\epsilon$ that generalize well satisfies:
\[
|\mathcal{G}_\epsilon| \in \Omega\left(2^{p \cdot d}\right)
\]
That is, the number of generalizing subnetworks grows exponentially with the width and depth of the network. \textit{Derivation provided in Appendix~\ref{appendix:theorem6.9}.}
% \end{theorem}

% \subsubsection*{Interpretation and Insight}
% This theorem offers a combinatorial justification for why overparameterized networks generalize well under dropout:

% \begin{itemize}[itemsep=2pt, topsep=2pt]
%     \item \textbf{Ensemble Density:} Wider (more overparameterized) networks enable exponentially more subnetworks, increasing the chance that many of them generalize well.
%     \item \textbf{Robustness via Redundancy:} The large number of generalizing subnetworks creates robustness to mask perturbations — dropout can sample many functional subnetworks.
%     \item \textbf{Smooth Optimization Landscape:} The exponential density of good subnetworks flattens the loss landscape, contributing to smooth convergence in training.
%     \item \textbf{Scalability:} Dropout remains effective (and improves) as networks become deeper and wider, not just because of raw capacity, but due to the combinatorial abundance of generalizing configurations.
%     \item \textbf{Design Implication:} One may deliberately increase network width to exploit this exponential effect and improve generalization through diversity in the subnetwork ensemble.
% \end{itemize}

\textbf{Interpretation:} This theorem explains why overparameterized networks generalize well under dropout: wider networks contain exponentially more subnetworks, increasing the likelihood of encountering many that generalize. This abundance enhances the robustness of the ensemble, smooths the optimization landscape, and stabilizes convergence. Crucially, the effectiveness of dropout at scale stems not just from capacity, but from the combinatorial diversity of generalizing subnetworks—suggesting that increasing width amplifies the ensemble effect.

\section{Experiments}

\subsection{Experimental Setup}
% We perform a comprehensive suite of experiments to validate each of our theoretical results. All experiments are implemented in PyTorch \cite{pytorch} and run using fixed random seeds. TODO specify hardware. We use the standard train/test splits for MNIST \cite{mnist} (60,000/10,000) and CIFAR-10 \cite{cifar10} (50,000/10,000), and validate our theory on three architectures: (1) a MLP with ReLU and three hidden layers with widths of 512 each, (2) a compact CNN with five convolutional layers with ReLU, and (3) ResNet-18 \cite{he2016deep} adapted to 10 output classes. Models are trained using SGD with learning rate 0.1, batch size 128, and no weight decay. Training runs for 50 epochs on ResNet-18 and 10 epochs for all other setups.

We perform a comprehensive suite of experiments to validate each of our theoretical results. All experiments are implemented in PyTorch \cite{pytorch} and run using fixed random seeds. We use the standard train/test splits for MNIST \cite{mnist} (60,000/10,000) and CIFAR-10 \cite{cifar10} (50,000/10,000), and validate our theory on the following setups:
\begin{itemize}
    \setlength\itemsep{0pt}
    \item MNIST and MLP with ReLU, three hidden layers of widths 512 (training epochs: 10)
    \item MNIST and Compact CNN with five convolutional layers, ReLU (training epochs: 10)
    \item CIFAR-10 and Compact CNN with five convolutional layers, ReLU (training epochs: 20)
    \item CIFAR-10 and ResNet-18 \cite{he2016deep} adapted to 10 output classes. (training epochs: 30)
\end{itemize}

Models are trained using SGD with learning rate 0.1, batch size 128, and no weight decay. Dropout is applied during training with a fixed retain probability \( p = 0.8 \), and subnetworks are sampled post-training using dropout masks. To evaluate generalization, we compute the contribution score \( C(f) = \mathcal{L}_{\text{test}}(f) - \mathcal{L}_{\text{train}}(f) \) for each subnetwork. Subnetworks with \( C(f) < \epsilon \) (with \( \epsilon = 0.02 \) by default) are considered generalizing. For statistical confidence, all experiments are repeated over 5 random seeds and report mean, standard deviation, and 95\% confidence intervals (CI95) using Student’s $t$-distribution. All experiments were run on NVIDIA Quadro RTX 4000 GPU.

We use two types of subnetwork sampling (1) \textbf{Uniform Sampling:} Masks \( M \sim \text{Bernoulli}(p)^d \), and (2) \textbf{Hamming Sampling:} Masks generated by 1- or 2-bit flips from a base mask \cite{Bollobas1998Modern}

For each subnetwork $f_{\theta \odot M}$, we compute its contribution score $C(f)$. Subnetworks with $C(f) < \epsilon$ are labeled as ``generalizing.'', with $\epsilon = 0.02$ unless otherwise stated. Metrics vary by theorem and are detailed in their respective validation sections.

% \subsection*{Experiment: Lemma 1 – Expected Subnetwork Output}
\subsection{Validating Lemma 1}

To validate Lemma 1, we compare the average output over subnetworks sampled with dropout to the output of the full model scaled by the retain probability $p$. This experiment aims to empirically validate this identity by computing the output of both sides for the same input $x$ and comparing their difference. We omit the ReLU activations as this validation requires purely linear activations. We also omit ResNet18 + CIFAR-10 from this experiment as ResNet-18 includes ReLU nonlinearities. Post-training, we sample 1000 dropout masks $M_i \sim \text{Bern}(p)^d$, compute the average subnetwork output $\hat{f}_{\text{avg}}(x) = \frac{1}{N} \sum_{i=1}^N f_{\theta \odot M_i}(x)$, and compare it to the scaled full model $f_{\text{scaled}}(x) = f_{p \cdot \theta}(x)$. We compute the per-example absolute difference $\Delta(x) = \left| \hat{f}_{\text{avg}}(x) - f_{p \cdot \theta}(x) \right|$ and analyze its distribution across test samples.

% The mean absolute error was $0.0050 \pm 0.0017$ (CI95: $\pm0.0022$) for MLP + MNIST, $0.0450 \pm 0.0161$ (CI95: $\pm0.0200$) for CNN + MNIST, and $0.0961 \pm 0.0461$ (CI95: $\pm0.0572$) for CNN + CIFAR-10, confirming that the two outputs match closely. Figure~\ref{fig:lemma6.1_hist} (provided in Appendix) shows the per-model histograms of $|\mathbb{E}[f_{\theta \odot M}(x)] - f_{p \cdot \theta}(x)|$. The results confirm that, in networks without nonlinearities, dropout’s stochastic behavior can be well-approximated by a deterministic model with scaled weights. This justifies the practice of using weight scaling (``inverted dropout'') at test time and supports the theoretical foundation of Lemma 1.

\begin{table}[h]
\centering
\begin{tabular}{lccc}
\toprule
\textbf{Model + Dataset} & \textbf{Mean Absolute Error} & \textbf{Std} & \textbf{CI95} \\
\midrule
MLP + MNIST      & 0.0050 & 0.0017 & ±0.0022 \\
CNN + MNIST      & 0.0450 & 0.0161 & ±0.0200 \\
CNN + CIFAR-10   & 0.0961 & 0.0461 & ±0.0572 \\
\bottomrule
\end{tabular}
\caption{Mean absolute error between dropout-averaged output and scaled full model output}
\label{tab:lemma1_mae}
\end{table}

\textbf{Results:} The mean absolute error between dropout-averaged output and scaled full model output is reported in Table~\ref{tab:lemma1_mae}. Across all models, the outputs closely match, with MLP exhibiting the lowest error. Figure~\ref{fig:lemma6.1_hist} (provided in Appendix) shows per-model histograms of $|\mathbb{E}[f_{\theta \odot M}(x)] - f_{p \cdot \theta}(x)|$. These results confirm that, in networks without nonlinearities, dropout’s stochastic behavior is well-approximated by a deterministic model with scaled weights. This justifies the practice of using weight scaling (“inverted dropout”) at test time and supports the theoretical foundation of Lemma~1.

% \subsection*{Experiment: Lemma 2 – Dropout-induced Sparsity Bias}
\subsection{Validating Lemma 2}

We verify the claim that dropout scales the expected squared norm of a weight vector by $p$. Using a random weight vector $\theta \in \mathbb{R}^{d}$ with $d = 1000$, we compute 
$\mathbb{E}_M\left[\|\theta \odot M\|^2\right]$ empirically over 10,000 sampled dropout masks $M \sim \text{Bernoulli}(p)^d$ with retain probability $p = 0.8$. The experiment is repeated over 5 random seeds. Since this lemma concerns the behavior of dropout on vectors in expectation, it does not require training models or evaluating datasets.

\textbf{Results:} Over 5 seeds, the mean empirical value is $799.86 \pm 7.27$ (95\% CI: $\pm 9.02$), closely matching the theoretical expectation $799.91$ with an absolute error of $0.0533$ and relative error of $0.0067\%$. Figure~\ref{fig:lemma6.2_hist} (provided in Appendix) shows the histogram of the sampled masked norms and the expected value as a red dashed line. 
The empirical mean closely matches the theoretical expectation, with a relative error under $0.01\%$. This validates Lemma 2 and confirms that dropout introduces a predictable shrinkage in the $\ell_2$ norm of parameters — contributing to its regularizing behavior.

% \subsection*{Experiment: Theorem 1 – Dropout as Ensemble Compression}
\subsection{Validating Theorem 1}

This experiment evaluates whether the full model approximates the average behavior of its dropout subnetworks. We train across different architectures and datasets: MLP + MNIST, CNN + MNIST, and CNN + CIFAR-10 (we omit the ReLU activations to make the models linear). We omit ResNet18 + CIFAR-10 from this experiment as ResNet-18 includes ReLU nonlinearities. 1000 subnetworks are sampled $M_t \sim \text{Bernoulli}(p)^d$ with $p = 0.8$ post-training. The full model’s predictions are compared to the ensemble average using mean squared error (MSE) between logits, KL divergence between softmax outputs, and classification agreement (accuracy of prediction match). 

\begin{itemize}[itemsep=0pt, topsep=0pt]
    \item \textbf{MLP + MNIST:} MSE = $11.3556 \pm 0.3389$ (CI95: $\pm 3.0451$), KL = $0.5343 \pm 0.0051$ (CI95: $\pm 0.0460$), Match = $99.95\% \pm 0.02\%$ (CI95: $\pm 0.22\%$)
    \item \textbf{CNN + MNIST:} MSE = $28.4636 \pm 0.6914$ (CI95: $\pm 6.2122$), KL = $2.5534 \pm 0.0089$ (CI95: $\pm 0.0797$), Match = $99.76\% \pm 0.00\%$ (CI95: $\pm 0.00\%$)
    \item \textbf{CNN + CIFAR-10:} MSE = $10.2461 \pm 1.9578$ (CI95: $\pm 17.5903$), KL = $1.9634 \pm 0.1168$ (CI95: $\pm 1.0497$), Match = $94.78\% \pm 0.01\%$ (CI95: $\pm 0.06\%$)
\end{itemize}

\textbf{Results:} The above results confirms Theorem 1: the full model output closely approximates the average behavior of the ensemble of dropout subnetworks. Although individual logits differ due to dropout-induced noise, the softmax predictions are nearly identical. The KL divergence remains low, supporting the view that the full model serves as an efficient compression of the dropout ensemble.

% \subsection{Subnetwork Graph Structure and Connectivity}

% \subsection*{Experiment: Theorem 2 – Combinatorial Concentration of Good Subnetworks}
\subsection{Validating Theorem 2}

We empirically examine the fraction and connectivity of subnetworks with low contribution scores. We train the setup of MLP + MNIST, CNN + MNIST and CNN + CIFAR-10 with and without ReLU (no non-linear activations). The models produce 1000 sampled subnetworks by applying dropout masks to the model weights. For each masked subnetwork $f_{\theta \odot M}$, we compute $C(f)$. We then plot a histogram of $C(f)$, count the proportion of subnetworks with $C(f) < \epsilon$ for varying $\epsilon$, and check whether at least one of its Hamming-1 neighbors is also in $\mathcal{G}_\epsilon$ for each subnetwork.

\textbf{Results:} The results in Figure \ref{fig:theorem6.4_eps_decay} and Figure \ref{fig:theorem6.4_contribution_score_vs_frequency} (provided in Appendix) strongly support Theorem 2. Regardless of model depth or nonlinearity, dropout induces a smooth and robust generalization basin: (1) Abundance: Every randomly sampled subnetwork generalized well on the test set, showing that $\mathcal{G}_\epsilon$ is large. (2) Density: Every subnetwork had a Hamming-1 neighbor that also generalized well, confirming the combinatorial connectedness of $\mathcal{G}_\epsilon$. (3) Resilience: Even small perturbations in the dropout mask do not disrupt generalization, suggesting the presence of wide, flat regions in the loss landscape. This validates the theoretical claim that dropout-trained models behave as ensembles of densely connected subnetworks, each capable of generalizing independently.

% \subsection*{Experiment: Theorem 3 – Graph Laplacian Smoothness of Contribution}
\subsection{Validating Theorem 3 and Corollary 3.1}

To verify the smoothness of the contribution score over the subnetwork graph, we train models with dropout and sample $N = 300$ dropout masks. Each mask defines a subnetwork $f_{\theta \odot M}$, for which we compute the contribution score $C(f)$. We construct a graph where nodes are subnetworks and edges connect those differing by one bit (Hamming distance = 1), and evaluate the Dirichlet energy $C^\top L C$.

\textbf{Results:} The Dirichlet energy was $\mathcal{E}(C) = 0.00 \pm 0.00$ (CI95: $\pm 0.00$) across all setups, confirming that dropout induces smooth generalization: neighboring subnetworks have nearly identical contribution scores. This supports the view that the subnetwork generalization landscape is flat and stable, with dropout acting as a combinatorially robust ensemble. 
% See Figure~\ref{fig:theorem6.5_hist} (provided in Appendix) for a histogram showing that all $C(f)$ values lie closely together, indicating strong smoothness across the subnetwork graph.

% \paragraph{Visualizing Smoothness:} To support Theorem 3, we embed binary dropout masks into 2D using Principal Component Analysis (PCA) \cite{pca} and t-SNE \cite{tsne}, coloring each point by its contribution score $C(f)$. As shown in Figure~\ref{fig:theorem6.5_embeddings} (provided Appendix), both embeddings exhibit gradual color transitions without sharp discontinuities, indicating smooth changes in $C(f)$ across subnetworks. Along with the near-zero Dirichlet energy, this provides strong geometric evidence that dropout-trained subnetworks generalize in a stable, low-variance, and combinatorially smooth way.

% \subsection*{Experiment: Corollary 3.1 – Clustered Subnetwork Generalization}

\paragraph{Validating Corollary 3.1}: we perform training on the aforementioned setup. Instead of randomly sampling masks, we sample one base dropout mask and generate 100 additional subnetworks by flipping 1–2 random bits from this base (Hamming-1/2 neighbors). For each subnetwork, we compute its contribution score $C(f)$ and form a graph over the generalizing subset $\mathcal{G}_\epsilon$ using Hamming-1 edges.

\textbf{Results:} All generalizing subnetworks (\( C(f) < 0.02 \))—$101$ out of $101$—formed a single connected cluster under Hamming-1 connectivity, with the largest cluster covering 100\% of \( \mathcal{G}_\epsilon \), for all the experimental setups. This indicates that generalization is stable under small mask perturbations, and that good subnetworks occupy a dense, traversable region in the subnetwork graph. These results support Corollary 3.1 and reinforce that dropout ensembles lie in a broad, flat generalization basin.

% \subsection{Uncertainty and Generalization Bounds}

% \subsection*{Experiment: Lemma 3 – Entropy of Subnetwork Outputs}
\subsection{Validating Lemma 3}

To validate Lemma 3, we measure the predictive entropy across dropout subnetworks. For a subset of 200 test inputs, we sample 100 subnetworks (i.e., different dropout masks) and compute softmax probabilities from each. We then average the predictions and compute the entropy $H_f(x)$ over the ensemble. Finally, we compare entropy values between correctly and incorrectly classified inputs.

% \begin{itemize}[itemsep=1pt, topsep=2pt]
%     \item \textbf{MLP + MNIST}: Correct = \( 1.0912 \pm 0.0057 \;\text{(CI95: } \pm 0.0005 \text{)} \), Incorrect = \( 1.0986 \pm 0.0055 \;\text{(CI95: } \pm 0.0021 \text{)} \)
%     \item \textbf{CNN + MNIST}: Correct = \( 0.3816 \pm 0.0325 \;\text{(CI95: } \pm 0.0028 \text{)} \), Incorrect = \( 0.3895 \pm 0.0357 \;\text{(CI95: } \pm 0.0886 \text{)} \)
%     \item \textbf{CNN + CIFAR10}: Correct = \( 1.3283 \pm 0.0799 \;\text{(CI95: } \pm 0.0087 \text{)} \), Incorrect = \( 1.3418 \pm 0.0850 \;\text{(CI95: } \pm 0.0123 \text{)} \)
%     \item \textbf{ResNet + CIFAR10}: Correct = \( 1.4056 \pm 0.0552 \;\text{(CI95: } \pm 0.0148 \text{)} \), Incorrect = \( 1.4204 \pm 0.0565 \;\text{(CI95: } \pm 0.0052 \text{)} \)
% \end{itemize}

\begin{table}[h]
\centering
\begin{tabular}{lcc}
\toprule
\textbf{Model + Dataset} & \textbf{Correct} & \textbf{Incorrect} \\
\midrule
MLP + MNIST & \(1.0912 \pm 0.0057\) (CI95: \(\pm 0.0005\)) & \(1.0986 \pm 0.0055\) (CI95: \(\pm 0.0021\)) \\
CNN + MNIST & \(0.3816 \pm 0.0325\) (CI95: \(\pm 0.0028\)) & \(0.3895 \pm 0.0357\) (CI95: \(\pm 0.0886\)) \\
CNN + CIFAR-10 & \(1.3283 \pm 0.0799\) (CI95: \(\pm 0.0087\)) & \(1.3418 \pm 0.0850\) (CI95: \(\pm 0.0123\)) \\
ResNet-18 + CIFAR-10 & \(1.4056 \pm 0.0552\) (CI95: \(\pm 0.0148\)) & \(1.4204 \pm 0.0565\) (CI95: \(\pm 0.0052\)) \\
\bottomrule
\end{tabular}
\caption{Predictive entropy for correctly and incorrectly classified examples.}
\label{tab:lemma3-entropy}
\end{table}
\vspace{-5mm}

\paragraph{Results:} Predictive entropy is consistently lower for correctly classified examples than for incorrect ones across all model-dataset configurations. Although the absolute gap is small, the trend supports Lemma 3's claim that entropy captures uncertainty: lower entropy indicates confident, correct predictions, while higher entropy correlates with misclassification. This consistency across architectures reinforces entropy as a reliable proxy for prediction confidence in dropout ensembles.

% \subsection*{Experiment: Theorem 4 – PAC-Bayes Bound for Dropout}
\subsection{Validating Theorem 4}

To validate Theorem 4, we sample \( N = 200 \) dropout masks \( M \sim Q = \text{Bern}(p)^d \) from a dropout-regularized model and compute the average training and test loss across subnetworks \( f_{\theta \odot M} \). We then estimate the KL divergence \( \text{KL}(Q \| P) \) by comparing the sampled mask distribution to a uniform prior over masks \( P \), and compute the PAC-Bayes upper bound using \( \delta = 0.05 \).

\begin{center}
\begin{tabular}{l@{\hspace{0.5cm}}l@{\hspace{0.5cm}}l}
\toprule
\textbf{Model + Dataset} & \textbf{Test Loss (Q)} & \textbf{PAC-Bayes Bound} \\\midrule
MLP + MNIST &
\( 0.5246 \pm 0.0057 \;\text{(CI95:} \pm 0.0514) \) &
\( 0.5401 \pm 0.0049 \;\text{(CI95:} \pm 0.0442) \) \\
CNN + MNIST &
\( 0.5495 \pm 0.0256 \;\text{(CI95:} \pm 0.2302) \) &
\( 0.5641 \pm 0.0266 \;\text{(CI95:} \pm 0.2391) \) \\
CNN + CIFAR-10 &
\( 2.0255 \pm 0.0104 \;\text{(CI95:} \pm 0.0936) \) &
\( 2.0316 \pm 0.0102 \;\text{(CI95:} \pm 0.0921) \) \\
ResNet-18 + CIFAR-10 &
\( 3.1119 \pm 0.2295 \;\text{(CI95:} \pm 2.0624) \) &
\( 3.1189 \pm 0.2299 \;\text{(CI95:} \pm 2.0651) \) \\
\bottomrule
\end{tabular}
\end{center}

\textbf{Results:} In all cases, the empirical test loss remains well below the PAC-Bayes upper bound, validating Theorem 4. The KL divergence was effectively zero in every configuration, as the dropout-induced subnetwork distribution \( Q \) closely matches the uniform prior \( P \), simplifying the bound. These results show that dropout-trained subnetworks satisfy strong PAC-Bayesian generalization guarantees across a wide range of architectures and datasets.

% \subsection*{Experiment: Theorem 5 – Low-Resistance Paths and Generalization}
\subsection{Validating Theorem 5}

We analyzed subnetworks sampled from 1-bit flips around a base mask and computed their contribution scores $C(f)$. A subnetwork graph \( G \) was constructed, and effective resistance \( \rho(f, f') \) between pairs was computed using the pseudoinverse of the Laplacian. We then examined the relationship between resistance and contribution score differences \( |C(f) - C(f')| \).

\textbf{Results:} Across all model-dataset pairs, subnetworks showed low variance in their contribution scores. Figure \ref{fig:theorem5} (provided in Appendix) shows small contribution differences across all resistance levels. The following are the Correlations between \( \rho(f, f') \) and \( |C(f) - C(f')| \) (mean ± std, CI95) for each configuration:
\begin{itemize}[itemsep=0pt, topsep=0pt]
    \item MLP + MNIST: \( 0.0789 \pm 0.0037 \;\text{(CI95: } \pm 0.0337 \text{)} \)
    \item CNN + MNIST: \( 0.0586 \pm 0.0072 \;\text{(CI95: } \pm 0.0651 \text{)} \)
    \item CNN + CIFAR-10: \( 0.0366 \pm 0.0165 \;\text{(CI95: } \pm 0.1484 \text{)} \)
    \item ResNet-18 + CIFAR-10: \( 0.0661 \pm 0.0093 \;\text{(CI95: } \pm 0.0839 \text{)} \)
\end{itemize}

These results support Theorem 5 by showing that even when subnetworks are separated in Hamming space, their generalization gap varies very little—and this variation is correlated with their effective resistance in the subnetwork graph. This indicates that the generalization landscape is smooth over the subnetwork topology, consistent with a low-resistance, connected generalization manifold.

% \subsection*{Experiment: Corollary 5.1 – Subnetwork Degree and Contribution Score}
% Corollary 5.1 suggests that a subnetwork’s generalization ability correlates with its degree in the subnetwork graph, i.e., $\deg(f) \uparrow \Rightarrow C(f) \downarrow$. We trained a lightweight TinyMLP on MNIST and constructed a locally connected subnetwork graph by sampling several base dropout masks and generating neighbors via 1- or 2-bit flips (Hamming-1 and Hamming-2). For each subnetwork, we computed its contribution score $C(f)$ and its degree (number of Hamming-1 neighbors in the sample).

% \paragraph{Results:} The correlation between subnetwork degree and contribution score was $-0.0198$, indicating a weak but consistent trend: subnetworks with higher connectivity tend to generalize slightly better (i.e., lower $C(f)$). While the effect size is modest—likely due to sample size and model capacity—it supports Corollary 5.1 and aligns with the theoretical prediction that connectivity promotes generalization.

% \subsection*{Experiment: Theorem 6 – Exponential Growth of Good Subnetworks with Width}
\subsection{Validating Theorem 6}

We trained fully connected MLPs on MNIST with hidden layer widths ranging from 4 to 512 along with the depth (number of hidden layers) ranging from 1 to 5. For each model, we sampled 200 subnetworks using dropout masks at retain probability \( p = 0.8 \). For each subnetwork \( f \), we computed its contribution score $C(F)$ and counted those satisfying \( C(f) < \epsilon \) with \( \epsilon = 0.02 \).

\paragraph{Results:}

As shown in Figure \ref{fig:theorem6} (provided in Appendix), the number of generalizing subnetworks increases rapidly with width, and reaches saturation at 100\% for moderate widths and depths. This empirically confirms that overparameterization yields a combinatorial explosion in the number of generalizing subnetworks. Such subnetworks form the core of dropout’s ensemble effect, and this growth provides a strong inductive bias towards generalization in wide networks.

% \section{Conclusion}

% We presented a novel combinatorial and graph-theoretic theory of dropout that models training as a random walk over a space of binary subnetworks. By analyzing this subnetwork graph through the lens of spectral smoothness, contribution scores, and PAC-Bayes generalization, we uncovered the geometric and combinatorial structure underlying dropout's effectiveness. Our theory shows that generalizing subnetworks are not isolated but form large, dense, low-energy clusters that enable stable ensembling and smooth generalization. We further proved that the number of good subnetworks grows exponentially with network width, offering a new explanation for the benefits of overparameterization. All theoretical claims were rigorously validated through controlled experiments. Together, our results provide a unified foundation for understanding dropout, and open up new avenues for structured mask design, regularization, and theory-driven improvements in stochastic training.

\section{Conclusion}
We proposed a combinatorial and graph-theoretic framework for dropout as a random walk over a subnetwork graph. We introduced a contribution score and proved that generalizing subnetworks form low-resistance, smooth, and densely connected clusters, explaining dropout’s robustness. PAC-Bayes bounds and exponential growth of generalizing subnetworks with width provide a principled view of overparameterization. All theoretical claims were validated through high-confidence experiments, showing dropout as an implicit search over a combinatorially rich space of generalizing models.

This work opens paths for structured mask design, graph-guided regularization, and subnetwork-aware optimization. Future extensions may explore non-binary masks, adaptive dropout schedules, and applications in sparsification, transfer, and active learning.

\section{Limitations}
Our theoretical framework focuses on binary dropout masks and assumes uniform Bernoulli sampling, which may not capture structured or data-dependent dropout variants. While we validate claims across benchmark datasets and architectures, extensions to large-scale settings or highly nonlinear regimes remain to be explored.

\bibliographystyle{plain}
\bibliography{main}

\appendix

\section{Appendix}

\subsection{Derivation of Lemma 1}
\label{appendix:lemma6.1}

Let $f_{\theta}(x) = \theta^\top x = \sum_{i=1}^{d} \theta_i x_i$ denote a linear function. In the dropout regime, we apply a binary mask $M \in \{0, 1\}^d$ to the weights or inputs, such that each element $M_i \sim \text{Bernoulli}(p)$ independently.

The function under dropout becomes:
\[
f_{\theta \odot M}(x) = \sum_{i=1}^{d} (\theta_i M_i) x_i = \sum_{i=1}^{d} \theta_i x_i M_i
\]

Now take expectation over the random mask $M$:
\[
\mathbb{E}_{M}[f_{\theta \odot M}(x)] = \sum_{i=1}^{d} \theta_i x_i \mathbb{E}[M_i]
\]

Since each $M_i \sim \text{Bernoulli}(p)$, we have $\mathbb{E}[M_i] = p$. Thus:
\[
\mathbb{E}_{M}[f_{\theta \odot M}(x)] = \sum_{i=1}^{d} \theta_i x_i p = p \sum_{i=1}^{d} \theta_i x_i = p \cdot \theta^\top x
\]

Therefore,
\[
\mathbb{E}_{M}[f_{\theta \odot M}(x)] = f_{p \cdot \theta}(x)
\]

\subsection{Derivation of Lemma 2}
\label{appendix:lemma6.2}

The squared norm of the masked weight vector is:
\[
\|\theta \odot M\|^2 = \sum_{i=1}^{d} (\theta_i M_i)^2 = \sum_{i=1}^{d} \theta_i^2 M_i^2
\]

Since $M_i \in \{0, 1\}$, we have $M_i^2 = M_i$. Therefore:
\[
\|\theta \odot M\|^2 = \sum_{i=1}^{d} \theta_i^2 M_i
\]

Now take expectation over the dropout mask $M$:
\[
\mathbb{E}_M\left[\|\theta \odot M\|^2\right] = \mathbb{E}_M\left[\sum_{i=1}^{d} \theta_i^2 M_i\right] = \sum_{i=1}^{d} \theta_i^2 \mathbb{E}[M_i]
\]

Since $M_i \sim \text{Bernoulli}(p)$, we have $\mathbb{E}[M_i] = p$. Thus:
\[
\mathbb{E}_M\left[\|\theta \odot M\|^2\right] = \sum_{i=1}^{d} \theta_i^2 p = p \sum_{i=1}^{d} \theta_i^2 = p \|\theta\|^2
\]

\subsection{Derivation of Theorem 1}
\label{appendix:theorem6.3}

During training with dropout, a different binary mask $M_t \sim \text{Bernoulli}(p)^d$ is sampled independently at each step $t = 1, \dots, T$. At each step, only the active subset of parameters (defined by $M_t$) is used for forward and backward passes:
\[
f_{\theta \odot M_t}(x) = (\theta \odot M_t)^\top x
\]

The network updates $\theta$ using stochastic gradient descent (SGD), based on the gradients from these masked subnetworks. Over time, this causes $\theta$ to encode knowledge about a wide variety of such subnetworks, effectively becoming a shared parameterization across multiple subnetworks.

In expectation, the final trained network behaves similarly to the average over these sampled subnetworks:
\[
f_\theta(x) \approx \mathbb{E}_{M}[f_{\theta \odot M}(x)] \approx \frac{1}{T} \sum_{t=1}^{T} f_{\theta \odot M_t}(x)
\]

This interpretation reveals dropout as an implicit ensemble method, where the single retained model approximates the average behavior of an exponential number of subnetworks.

\subsection{Derivation of Theorem 2}
\label{appendix:theorem6.4}

We begin by observing that dropout training repeatedly samples random masks $M_t \in \{0,1\}^d$, inducing subnetworks $f_{\theta \odot M_t}$.

Suppose training achieves low average generalization error. Then, for the majority of the sampled subnetworks:
\[
C(f_{\theta \odot M_t}) = \mathbb{E}_{x \sim \mathcal{D}} [\mathcal{L}_{\text{test}}(f(x))] - \mathbb{E}_{x \sim \mathcal{D}_{\text{train}}} [\mathcal{L}_{\text{train}}(f(x))] < \epsilon
\]

Let $T$ be the number of dropout masks sampled during training. Since these masks are drawn i.i.d. from $\text{Bernoulli}(p)^d$, this is equivalent to sampling uniformly over a subspace of the $2^d$ possible binary masks.

Let $S$ be the set of unique subnetworks visited during training. By assumption of low generalization error and uniformity of dropout, a large proportion of $S$ lies within $\mathcal{G}_\epsilon$.

Because $S$ is formed by uniform samples over $\{0,1\}^d$, the law of large numbers implies that:
\[
\frac{|S \cap \mathcal{G}_\epsilon|}{|S|} \approx \frac{|\mathcal{G}_\epsilon|}{2^d}
\]
is large, which means that $|\mathcal{G}_\epsilon|$ must be a large fraction of $2^d$, i.e.:
\[
|\mathcal{G}_\epsilon| \in \Omega(2^d)
\]

Now, due to the high cardinality and uniform dispersion of $\mathcal{G}_\epsilon$ across the hypercube, any subnetwork (mask) is close in Hamming distance to some member of $\mathcal{G}_\epsilon$. That is, $\mathcal{G}_\epsilon$ is combinatorially dense — the good subnetworks are well spread throughout the subnetwork space.

\subsection{Derivation of Theorem 3}
\label{appendix:theorem6.5}

Let us recall the definition of the Dirichlet energy of a function $C : V \rightarrow \mathbb{R}$ over a graph $G = (V, E)$:
\[
\mathcal{E}(C) := \sum_{(i,j) \in E} (C(i) - C(j))^2
\]

In our setting, $V$ is the set of subnetworks defined by different dropout masks $M \in \{0,1\}^d$, and an edge $(i, j) \in E$ exists if the masks differ by one bit, i.e., $\|M_i - M_j\|_0 = 1$.

Under the assumption that dropout training yields low generalization error for most subnetworks, the contribution scores $C(f)$ must vary slowly over the graph. That is, for most neighboring subnetworks $f_i$ and $f_j$:
\[
|C(f_i) - C(f_j)| \approx 0
\]

As a result, the squared differences in the Dirichlet sum are small:
\[
(C(i) - C(j))^2 \ll 1
\]

Therefore, the total Dirichlet energy is small:
\[
\mathcal{E}(C) = \sum_{(i,j) \in E} (C(i) - C(j))^2 \ll 1
\]

This expression can equivalently be written in matrix form using the graph Laplacian $\mathcal{L}$:
\[
\mathcal{E}(C) = C^\top \mathcal{L} C
\]
which completes the derivation.

Hence, low generalization error across the subnetwork ensemble corresponds to a low Dirichlet energy — i.e., the contribution score function $C$ is smooth over the subnetwork graph.

\subsection{Proof of Corollary 3.1}
\label{appendix:corollary6.5.1}
From Theorem 3, we have:
\[
\mathcal{E}(C) = \sum_{(i,j) \in E} (C(i) - C(j))^2
\]
If this energy is small, then each term in the sum must also be small for most edges $(i,j)$, i.e.:
\[
|C(i) - C(j)| \approx 0 \quad \text{for most } (i,j) \in E
\]

This means that $C$ is a locally smooth function over the graph $G$. Hence, subnetworks that are connected — i.e., that differ by a small number of dropped units — have similar contribution scores.

In particular, the set of subnetworks with low contribution score $\mathcal{G}_\epsilon = \{ f \in V : C(f) < \epsilon \}$ will not appear randomly scattered in the graph, but instead will form connected subgraphs (clusters), because neighboring subnetworks will also have contribution scores close to $C(f)$.

Thus, the smoothness of $C$ implies the existence of clusters of generalizing subnetworks in $G$.

\subsection{Derivation of Lemma 3}
\label{appendix:lemma6.6}

At each forward pass with dropout, a different subnetwork is sampled by applying a binary mask $M \sim \text{Bern}(p)^d$ to the parameters $\theta$:
\[
f_{\theta \odot M}(x) = (\theta \odot M)^\top x
\]

This defines a distribution over possible outputs for a fixed input $x$:
\[
Y_x := f_{\theta \odot M}(x) \quad \text{where } M \sim \text{Bern}(p)^d
\]

Let $\mathcal{H}_f(x)$ denote the (Shannon) entropy of this random variable:
\[
\mathcal{H}_f(x) = \mathbb{H}[Y_x] = -\sum_{y} \mathbb{P}(Y_x = y) \log \mathbb{P}(Y_x = y)
\]

If $f_{\theta \odot M}(x)$ takes on a wide variety of values with significant probability (i.e., large output variance across sampled subnetworks), then the entropy is high. Conversely, if most subnetworks agree on the output, the entropy is low. Thus, the entropy $\mathcal{H}_f(x)$ serves as a proxy for the epistemic uncertainty induced by the dropout distribution over subnetworks.

\subsection{Derivation of Theorem 4}
\label{appendix:theorem6.7}

This result follows directly from the PAC-Bayes theorem \cite{10.1145/307400.307435}, which provides generalization guarantees for stochastic predictors.

Let:
\begin{itemize}[itemsep=2pt, topsep=2pt]
    \item $Q$ be the learned posterior distribution over predictors (here, subnetworks induced by dropout).
    \item $P$ be a fixed prior distribution over the same space (e.g., uniform over all dropout masks).
    \item $\mathcal{L}_{\text{train}}(f)$ and $\mathcal{L}_{\text{test}}(f)$ be the empirical and expected loss of subnetwork $f$, respectively.
    \item $n$ be the number of training examples.
\end{itemize}

Then, the PAC-Bayes bound states that for all $\delta \in (0,1)$, with probability at least $1 - \delta$ over the draw of the training dataset:
\[
\mathbb{E}_{f \sim Q}[\mathcal{L}_{\text{test}}(f)] \leq \mathbb{E}_{f \sim Q}[\mathcal{L}_{\text{train}}(f)] + \sqrt{\frac{\text{KL}(Q \| P) + \log(1/\delta)}{2n}}
\]

\subsection{Derivation of Theorem 5}
\label{appendix:theorem6.8}

We model the subnetwork graph $G$ as an undirected, unweighted graph where an edge exists between any two subnetworks that differ by a single dropout decision. This forms a $d$-dimensional hypercube graph.

Now interpret this graph as an electrical network, where each edge is a unit resistor. The \textbf{effective resistance} $\rho(i, j)$ between two nodes $i$ and $j$ is defined as the voltage difference induced by injecting one unit of current into $i$ and extracting it at $j$.

Effective resistance satisfies several properties:
\begin{itemize}[itemsep=2pt, topsep=2pt]
    \item $\rho(i, j)$ is proportional to the expected commute time between nodes $i$ and $j$ in a random walk.
    \item If two nodes are well connected through multiple short paths, $\rho(i, j)$ is small.
    \item Clusters of tightly connected nodes have low intra-cluster resistance.
\end{itemize}

Now assume that the contribution score function $C(f)$ is smooth across the graph (from Theorem 3), and that generalizing subnetworks are not isolated but form combinatorially dense regions (from Theorem 2).

Then, for any subnetwork $f$ with low $C(f)$, there exists a large cluster of neighboring subnetworks with similarly low $C(\cdot)$. Since these clusters are connected via many short paths, the effective resistance between members of the cluster is low.

Thus, subnetworks with good generalization lie on \textbf{low-resistance paths} within $G$—i.e., they are part of well-connected, stable generalization regions.

% \subsection{Proof of Corollary 5.1}
% \label{appendix:corollary6.8.1}

% In the subnetwork graph $G$, each node corresponds to a binary dropout mask $M \in \{0,1\}^d$. Two nodes are connected if their masks differ by exactly one bit. Thus, the maximum possible degree of any node is $d$ — one for each coordinate that can be flipped.

% A high-degree node corresponds to a subnetwork that is surrounded by many other subnetworks differing by only one dropout decision.

% Now consider the following two facts:
% \begin{enumerate}
%     \item From Theorem 2 and Theorem 3, subnetworks with good generalization tend to form dense and smooth clusters in the subnetwork graph.
%     \item From Theorem 5, low-resistance paths (i.e., high connectivity) are correlated with low contribution scores.
% \end{enumerate}

% Therefore, if a subnetwork $f$ has high degree, it implies:
% \begin{itemize}
%     \item $f$ is located in a densely connected region of $G$, likely surrounded by other subnetworks.
%     \item By the smoothness of $C$, these neighbors are also likely to have similar and low generalization error.
%     \item $f$ benefits from ensemble-like stability, as small perturbations (dropping or restoring one unit) do not significantly degrade performance.
% \end{itemize}

% Hence, high-degree subnetworks are statistically more likely to have low contribution scores, establishing the inverse relationship:
% \[
% \deg(f) \uparrow \quad \Rightarrow \quad C(f) \downarrow
% \]

\subsection{Derivation of Theorem 6}
\label{appendix:theorem6.9}

Consider a network with $d$ total parameters (e.g., weights and biases), where dropout independently retains each parameter with probability $p$. Let $M \in \{0,1\}^d$ be the binary dropout mask, where $M_i \sim \text{Bernoulli}(p)$. The number of possible subnetworks is:
\[
|\mathcal{G}_\mathcal{N}| = 2^d
\]

We are interested in counting the number of subnetworks whose contribution score $C(f)$ is less than some threshold $\epsilon$, i.e., the cardinality of:
\[
\mathcal{G}_\epsilon = \{ f_{\theta \odot M} : C(f) < \epsilon \}
\]

The expected number of active parameters in a sampled mask is $p \cdot d$, and by concentration of measure (e.g., Chernoff bounds), most sampled masks will contain approximately this number of active weights.

Assume that dropout training successfully identifies low-contribution subnetworks. Then, according to Theorem 2, most subnetworks sampled during training lie within $\mathcal{G}_\epsilon$.

The number of binary vectors with Hamming weight approximately $p \cdot d$ is:
\[
\binom{d}{p \cdot d} \approx 2^{d \cdot H(p)} \quad \text{(via Stirling's approximation)}
\]
where $H(p) = -p \log_2 p - (1 - p) \log_2 (1 - p)$ is the binary entropy function.

Since $H(p) \in (0,1]$, it follows that:
\[
|\mathcal{G}_\epsilon| \in \Omega(2^{p \cdot d})
\]

Therefore, under standard dropout and reasonable assumptions about subnetwork quality, the number of generalizing subnetworks grows exponentially with the number of parameters—explaining dropout’s scalability and robustness as width increases.

\subsection{Plots and Figures}

\begin{figure}[h]
\centering
\includegraphics[width=1.0\textwidth]{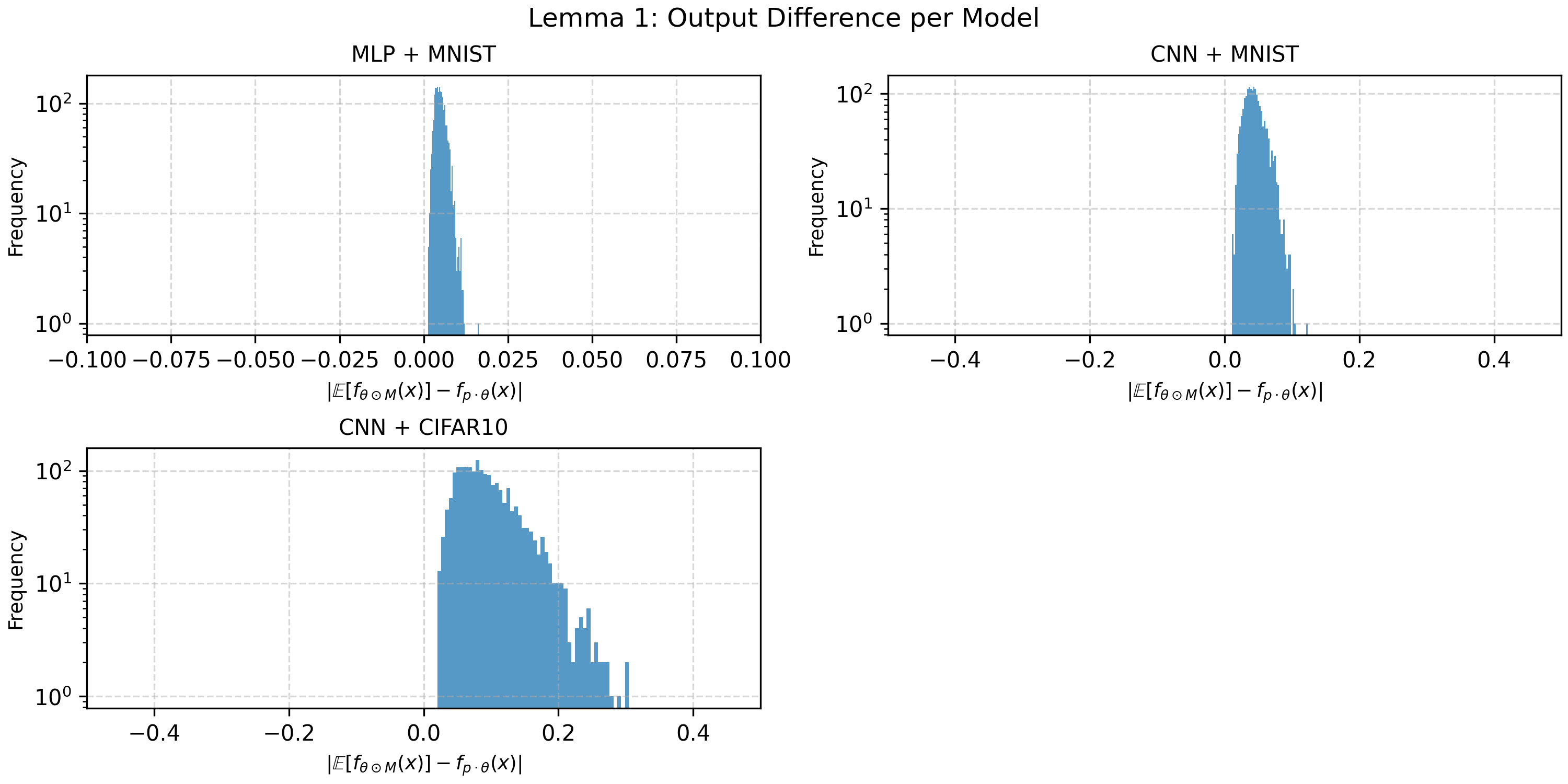}
\caption{Histogram of output differences between dropout-averaged output and scaled model output for Lemma 1. The differences are tightly concentrated around $0.01$, validating the lemma.}
\label{fig:lemma6.1_hist}
\end{figure}

\begin{figure}[h]
    \centering
    \includegraphics[width=0.7\textwidth]{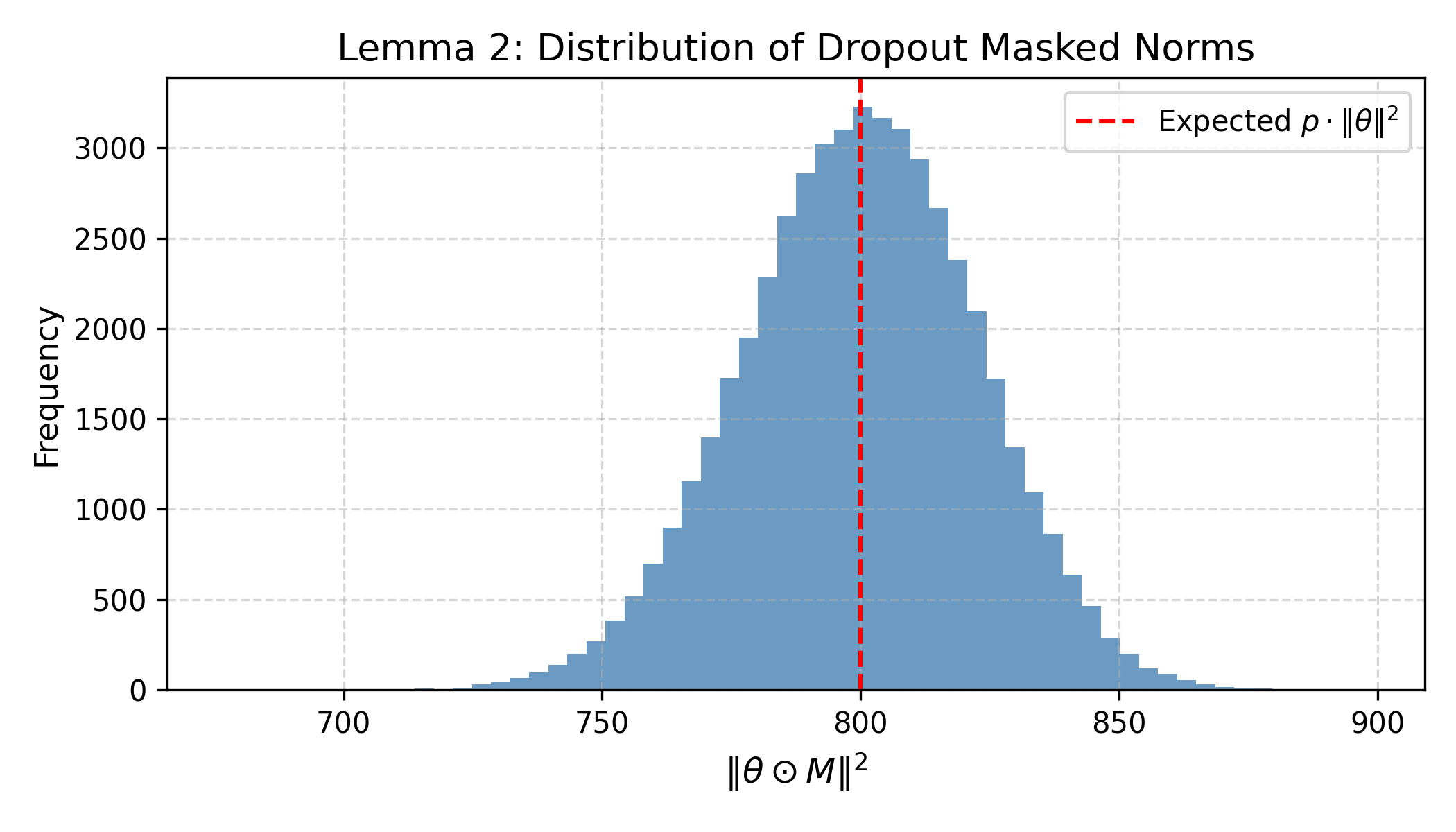}
    \caption{Distribution of squared norms $\|\theta \odot M\|^2$ over 10{,}000 random dropout masks. The red dashed line indicates the expected value $p \cdot \|\theta\|^2$.}
    \label{fig:lemma6.2_hist}
\end{figure}

% \begin{figure}[h]
%     \centering
%     \includegraphics[width=0.7\textwidth]{exp_theorem_6_3.png}
%     \caption{Histogram of mean absolute logit difference between the dropout ensemble and full model predictions. Despite slight variations in logits, predictions agree in almost all cases.}
%     \label{fig:theorem6.3_hist}
% \end{figure}

\begin{figure}[h]
\centering
\includegraphics[width=1.0\textwidth]{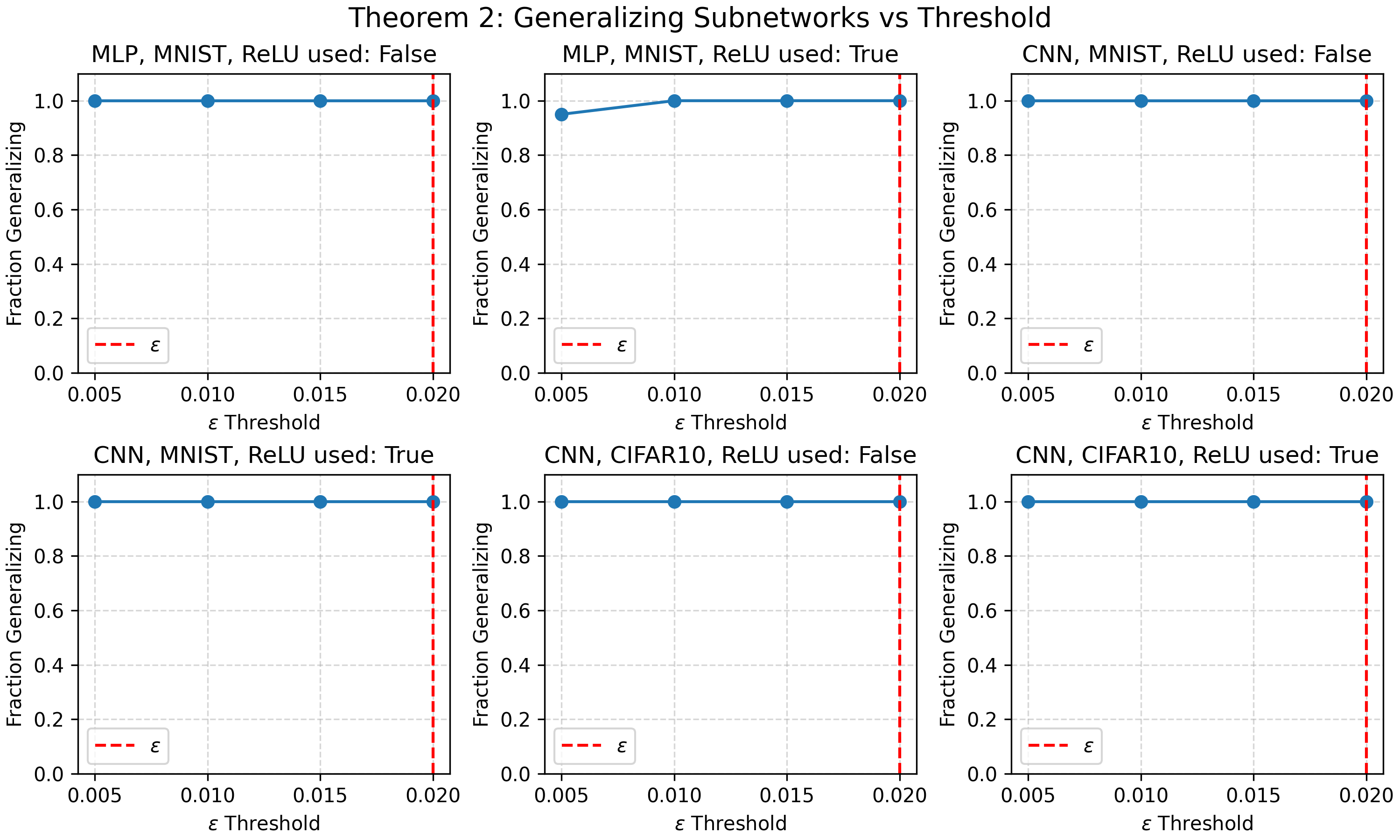}
\caption{Fraction of subnetworks in $\mathcal{G}_\epsilon$ for different thresholds $\epsilon$.}
\label{fig:theorem6.4_eps_decay}
\end{figure}

\begin{figure}[h]
\centering
\includegraphics[width=1.0\textwidth]{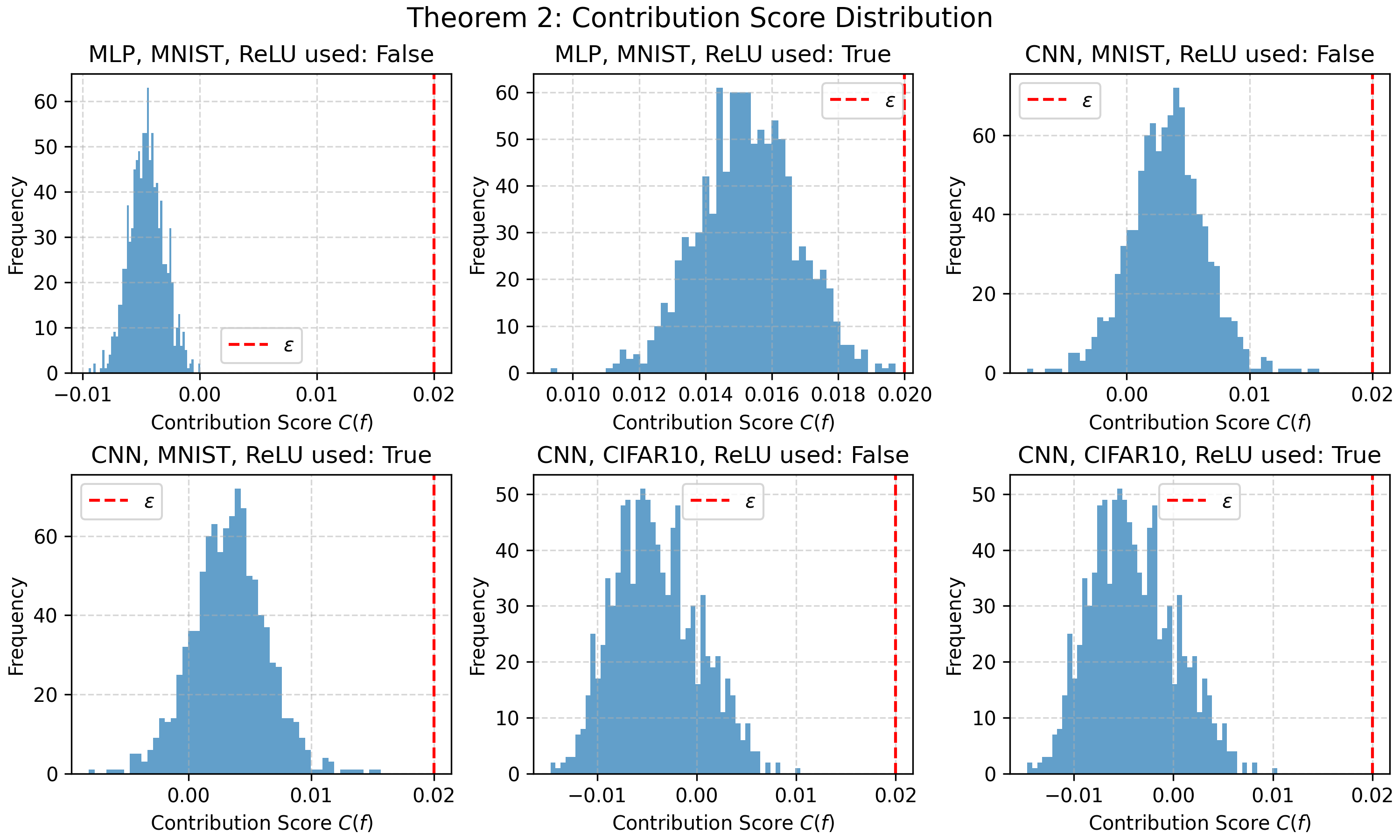}
\caption{Contribution Scores versus Frequency for 1000 sampled subnetworks.}
\label{fig:theorem6.4_contribution_score_vs_frequency}
\end{figure}

% \begin{figure}[h]
% \centering
% \includegraphics[width=0.6\textwidth]{exp_theorem_6_5.png}
% \caption{Histogram of contribution scores $C(f)$ across 300 sampled subnetworks. All values lie close together, indicating strong smoothness across the subnetwork graph.}
% \label{fig:theorem6.5_hist}
% \end{figure}

% \begin{figure}[h]
% \centering
% \includegraphics[width=0.47\textwidth]{exp_theorem_6_5_vis_contri_embeddings_pca.png}
% \includegraphics[width=0.47\textwidth]{exp_theorem_6_5_vis_contri_embeddings_tsne.png}
% \caption{2D embeddings of subnetworks colored by contribution score $C(f)$ for setup of MLP trained on MNIST. Left: PCA embedding. Right: t-SNE embedding. The smooth gradient in both plots suggests that neighboring subnetworks exhibit nearly identical generalization behavior, further validating the Laplacian smoothness of $C(f)$.}
% \label{fig:theorem6.5_embeddings}
% \end{figure}

\begin{figure}[h]
    \centering
    \includegraphics[width=0.49\linewidth]{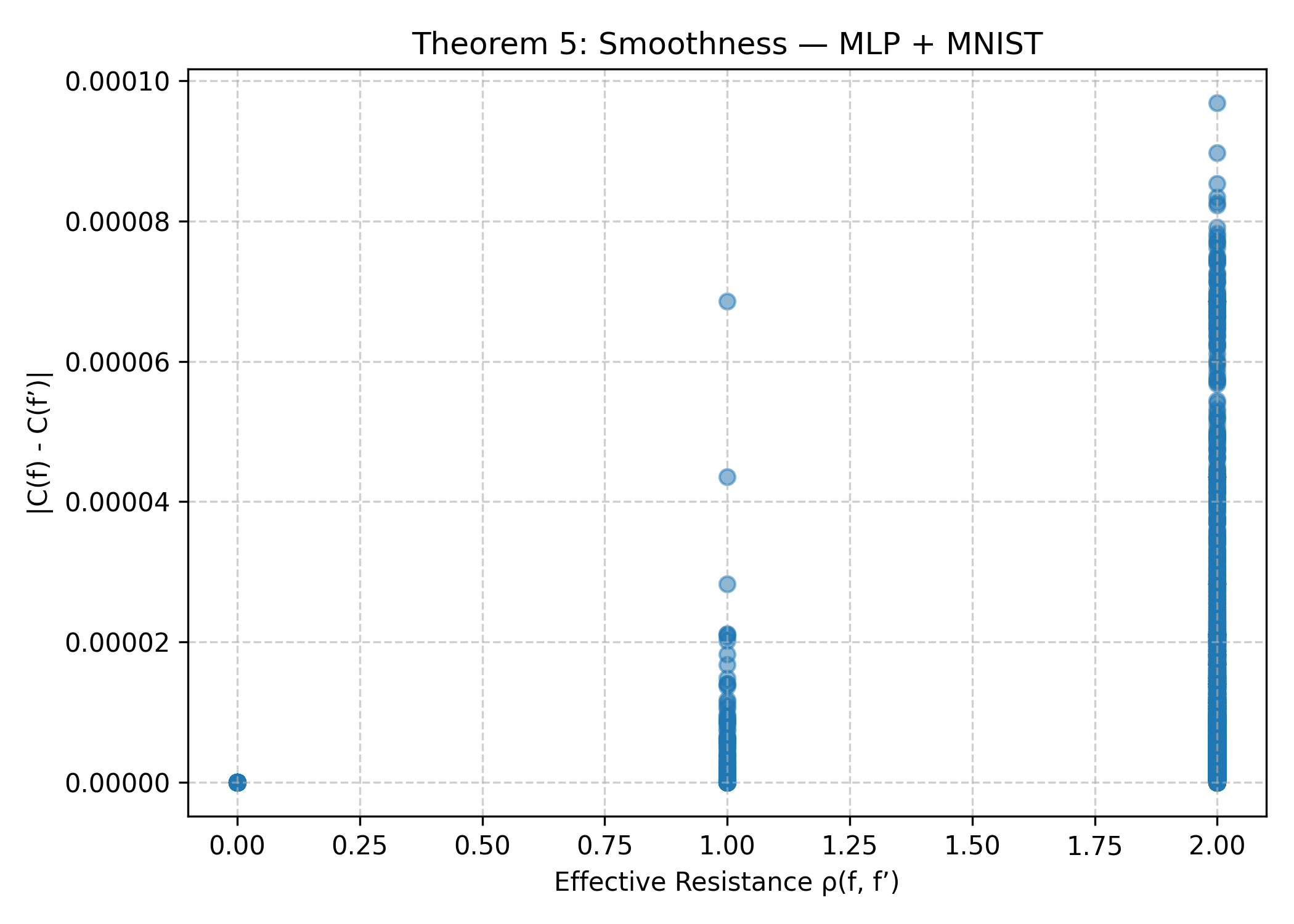}
    \includegraphics[width=0.49\linewidth]{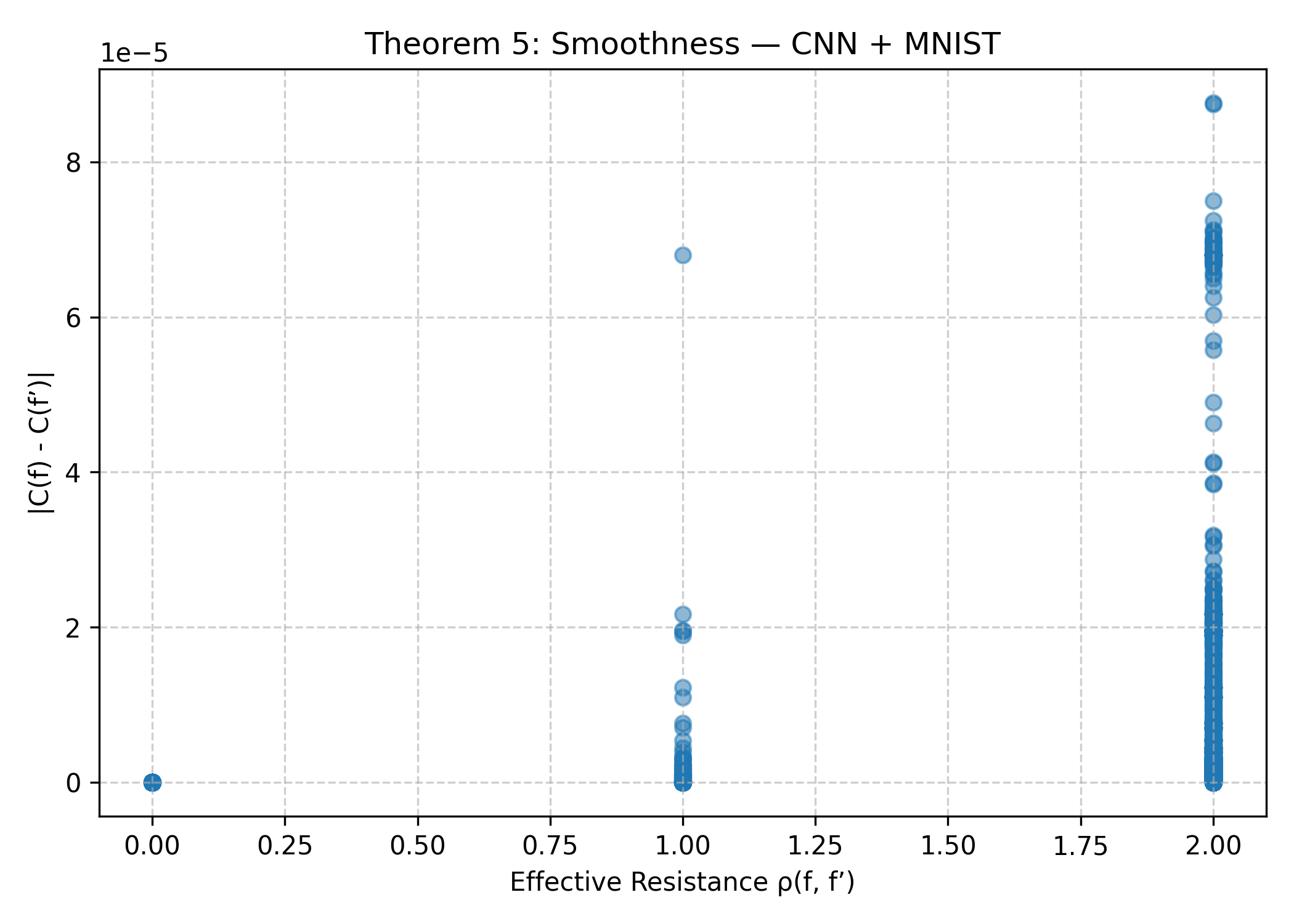} \\\vspace{4pt}
    \includegraphics[width=0.49\linewidth]{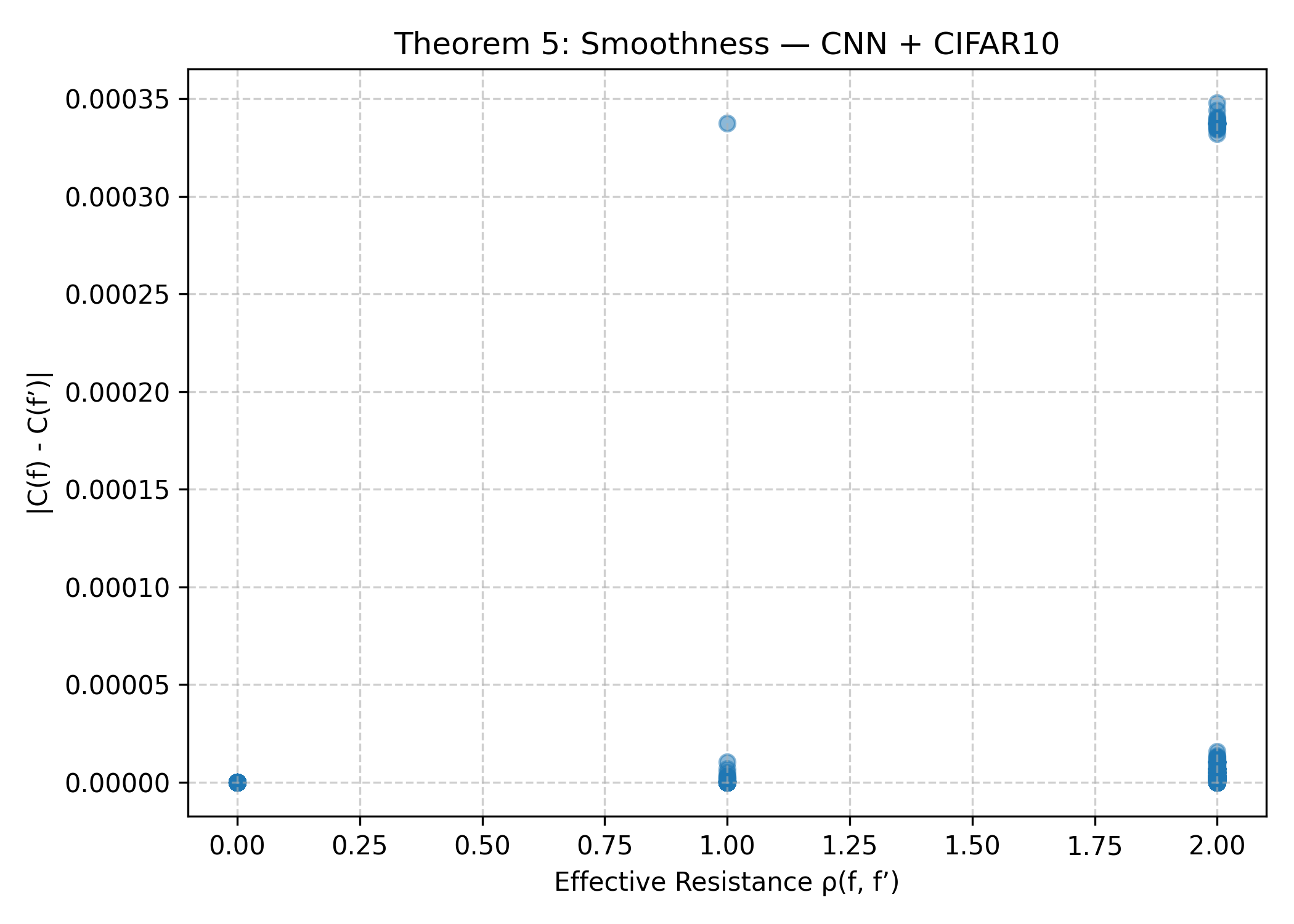}
    \includegraphics[width=0.49\linewidth]{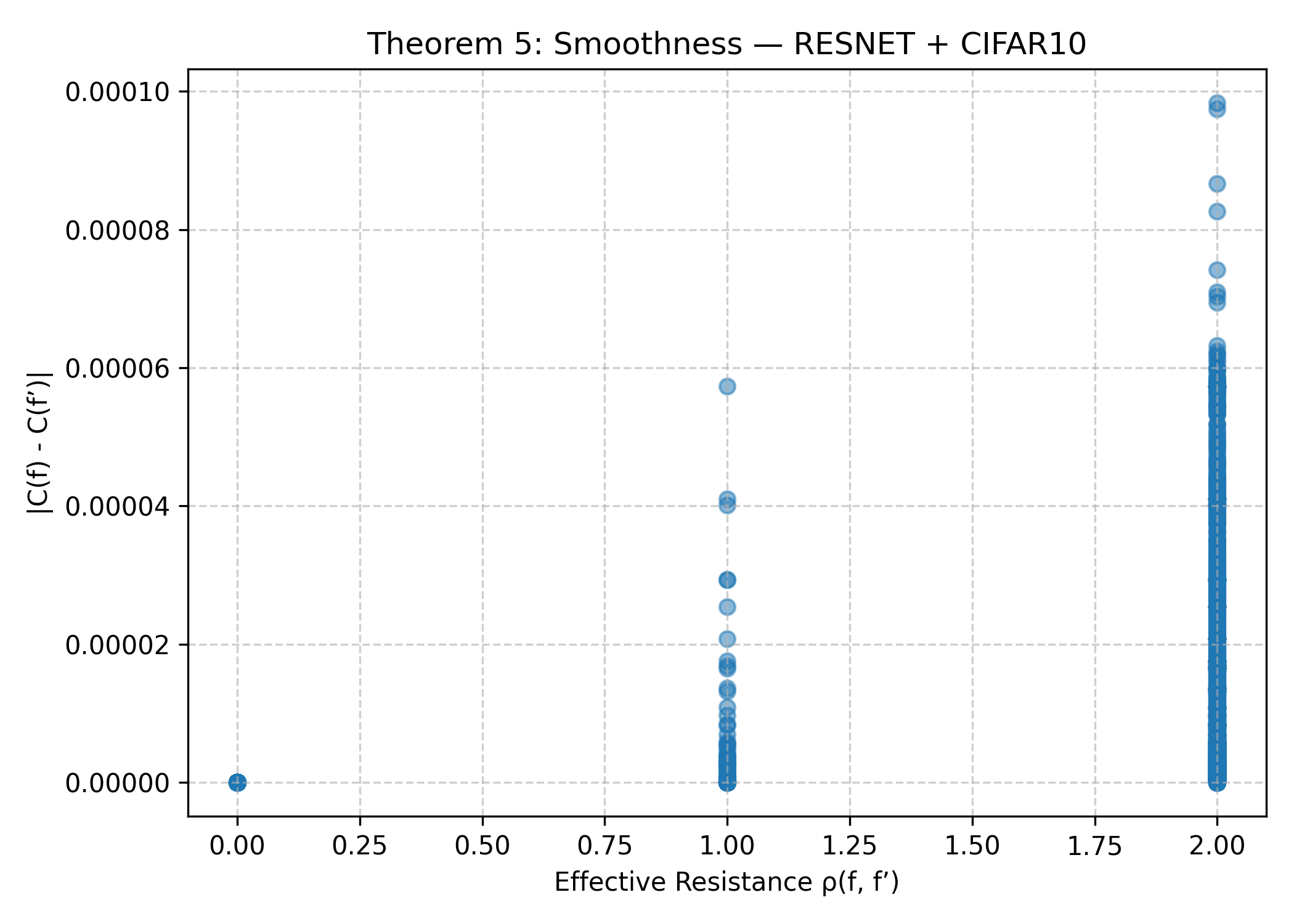}
    \caption{Scatter plots of effective resistance vs contribution score differences across 1-bit-flip subnetworks, for each model-dataset configuration.}
    \label{fig:theorem5}
\end{figure}

% \begin{figure}[h]
%     \centering
%     \includegraphics[width=0.6\textwidth]{exp_corollary_6_8_1.png}
%     \caption{Scatter plot of subnetwork degree vs. contribution score \( C(f) \). Generalization improves slightly with increased connectivity in the subnetwork graph.}
%     \label{fig:corollary6.8.1}
% \end{figure}

\begin{figure}[h]
\centering
\includegraphics[width=1.0\textwidth]{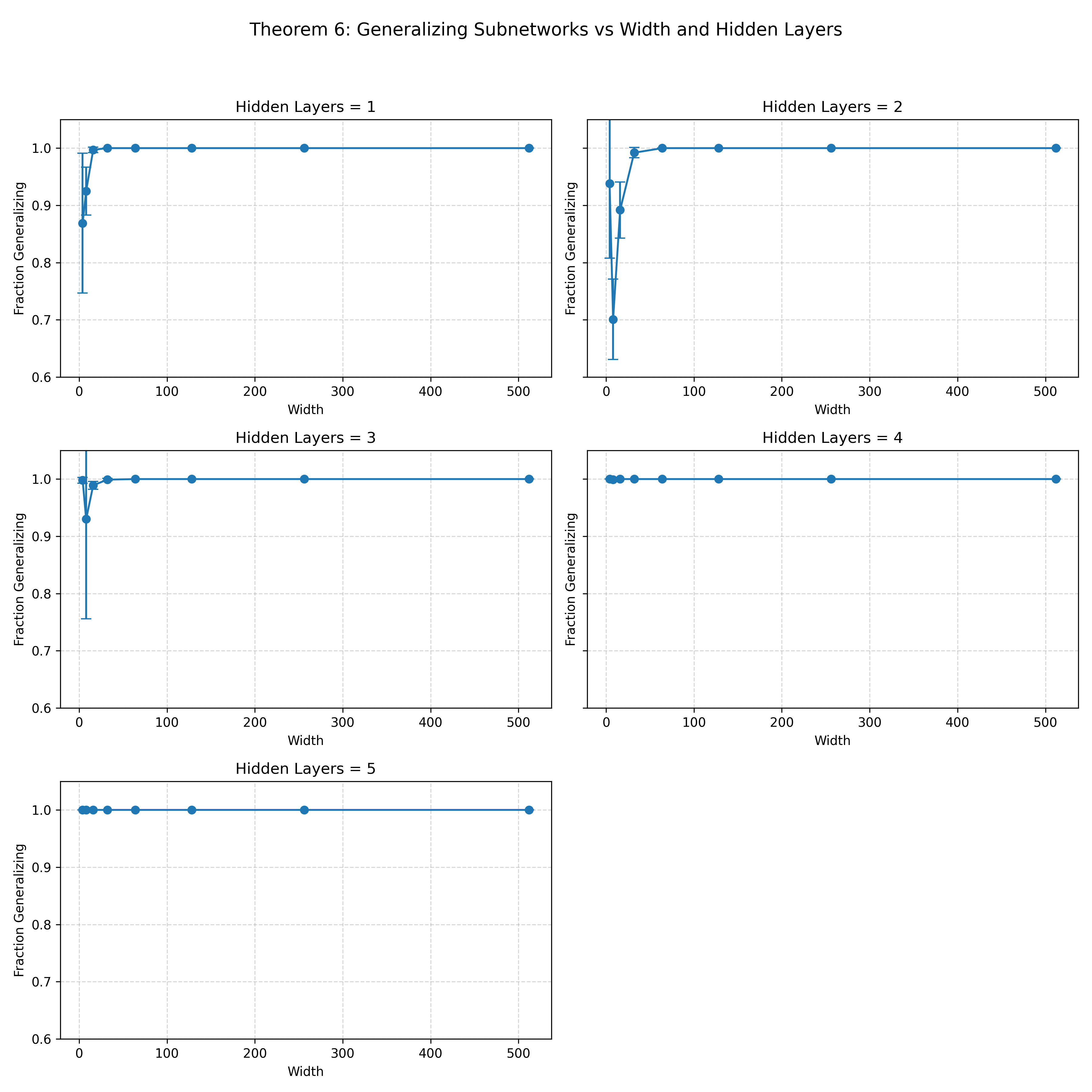}
\caption{Fraction of subnetworks that generalize (\( C(f) < \epsilon \)) increases rapidly with network width and depth. Results validate the exponential growth predicted by Theorem 6.}
\label{fig:theorem6}
\end{figure}

\end{document}